%% file: tomm.tex
\documentclass[manuscript]{acmart}

\usepackage{amsmath}
\usepackage{multirow}
\usepackage{booktabs} 
\usepackage{caption}
\usepackage{subcaption}
\usepackage{units} 

\newcommand{\figref}[1]{\mbox{Fig.~\ref{#1}}}
\newcommand{\tabref}[1]{\mbox{Table~\ref{#1}}}
\newcommand{\eqnref}[1]{\mbox{Eqn.~\ref{#1}}}

\newcommand{\etal}{\textit{et al}.}
\newcommand{\ie}{\textit{i}.\textit{e}.,}
\newcommand{\eg}{\textit{e}.\textit{g}.,}

\AtBeginDocument{%
  \providecommand\BibTeX{{%
    \normalfont B\kern-0.5em{\scshape i\kern-0.25em b}\kern-0.8em\TeX}}}

\setcopyright{acmcopyright}
\copyrightyear{2022}
\acmYear{2022}
\acmDOI{10.1145/1122445.1122456}

\acmConference[Woodstock '18]{Woodstock '18: ACM Symposium on Neural
  Gaze Detection}{June 03--05, 2018}{Woodstock, NY}
\acmBooktitle{Woodstock '18: ACM Symposium on Neural Gaze Detection,
  June 03--05, 2018, Woodstock, NY}
\acmPrice{15.00}
\acmISBN{978-1-4503-XXXX-X/18/06}

\begin{document}

\title{On Modality Bias Recognition and Reduction}


\author{Yangyang Guo}
\orcid{0000-0001-8691-5372}
\affiliation{
    \institution{National University of Singapore}
    \country{Singapore}}
\email{guoyang.eric@gmail.com}

\author{Liqiang Nie}
\orcid{0000-0003-1476-0273}
\affiliation{
    \institution{Harbin Institute of Technology (Shenzhen)}
    \country{China}}
\email{nieliqiang@gmail.com}

\author{Harry Cheng}
\orcid{0000-0001-7436-0162}
\affiliation{
    \institution{Shandong University}
    \country{China}}
\email{xaCheng1996@gmail.com}

\author{Zhiyong Cheng}
\orcid{0000-0003-1109-5028}
\affiliation{
    \institution{Shandong Artificial Intelligence Institute}
    \country{China}}
\email{jason.zy.cheng@gmail.com}

\author{Mohan Kankanhalli}
\orcid{0000-0002-4846-2015}
\affiliation{
    \institution{National University of Singapore}
    \country{Singapore}}
\email{mohan@comp.nus.edu.sg}

\author{Alberto Del Bimbo}
\orcid{0000-0002-1052-8322}
\affiliation{
    \institution{University of Florence}
    \country{Italy}}
\email{alberto.delbimbo@unifi.it}


\renewcommand{\shortauthors}{Yangyang Guo, et al.}

\begin{abstract}
Making each modality in multi-modal data contribute is of vital importance to learning a versatile multi-modal model. Existing methods, however, are often dominated by one or few of modalities during model training, resulting in sub-optimal performance. In this paper, we refer to this problem as modality bias and attempt to study it in the context of multi-modal classification systematically and comprehensively. After stepping into several empirical analyses, we recognize that one modality affects the model prediction more just because this modality has a spurious correlation with instance labels. In order to primarily facilitate the evaluation on the modality bias problem, we construct two datasets respectively for the colored digit recognition and video action recognition tasks in line with the Out-of-Distribution (OoD) protocol. Collaborating with the benchmarks in the visual question answering task, we empirically justify the performance degradation of the existing methods on these OoD datasets, which serves as evidence to justify the modality bias learning. In addition, to overcome this problem, we propose a  plug-and-play loss function method, whereby the feature space for each label is adaptively learned according to the training set statistics. Thereafter, we apply this method on ten baselines in total to test its effectiveness. From the results on four datasets regarding the above three tasks, our method yields remarkable performance improvements compared with the baselines, demonstrating its superiority on reducing the modality bias problem.
\end{abstract}



\maketitle

\input{segments/introduction}
\input{segments/related_work}
\input{segments/definition}
\input{segments/method}
\input{segments/experimental_set}
\input{segments/experimental_results}
\input{segments/conclusion}
\input{segments/appendix}

\begin{acks}
This research is supported by the National Research Foundation, Singapore under its Strategic Capability Research Centres Funding Initiative. Any opinions, findings and conclusions or recommendations expressed in this material are those of the author(s) and do not reflect the views of National Research Foundation, Singapore.
\end{acks}

\bibliographystyle{reference/ACM-Reference-Format}
\bibliography{tomm}



\end{document}

%% file: segments/introduction.tex
\section{Introduction}\label{introduction}
Real-world data often exhibit multiple modalities. Learning a versatile multi-modal model has naturally attracted increasing research interests from academic and industrial practitioners. Existing cutting edge technologies mostly resort to multi-modal machine learning or deep learning~\cite{addition, captiontomm}, wherein considerable advancement has been brought by the computer vision and natural language processing communities. Despite promising progress having been achieved, we recognize one detrimental problem in multi-modal learning, which severely degrades the decision making integrity. That is, current models prefer trivial solutions due to the shortcut between targets and certain modalities. This phenomenon is however, ubiquitous among various multi-modal learning tasks, including classification, generation and clustering. In this work, we shed light on the context of multi-modal classification. It takes as input the mixed modality data and outputs a human-friendly semantic label.

\begin{figure}
  \centering
  \includegraphics[width=1.0\linewidth]{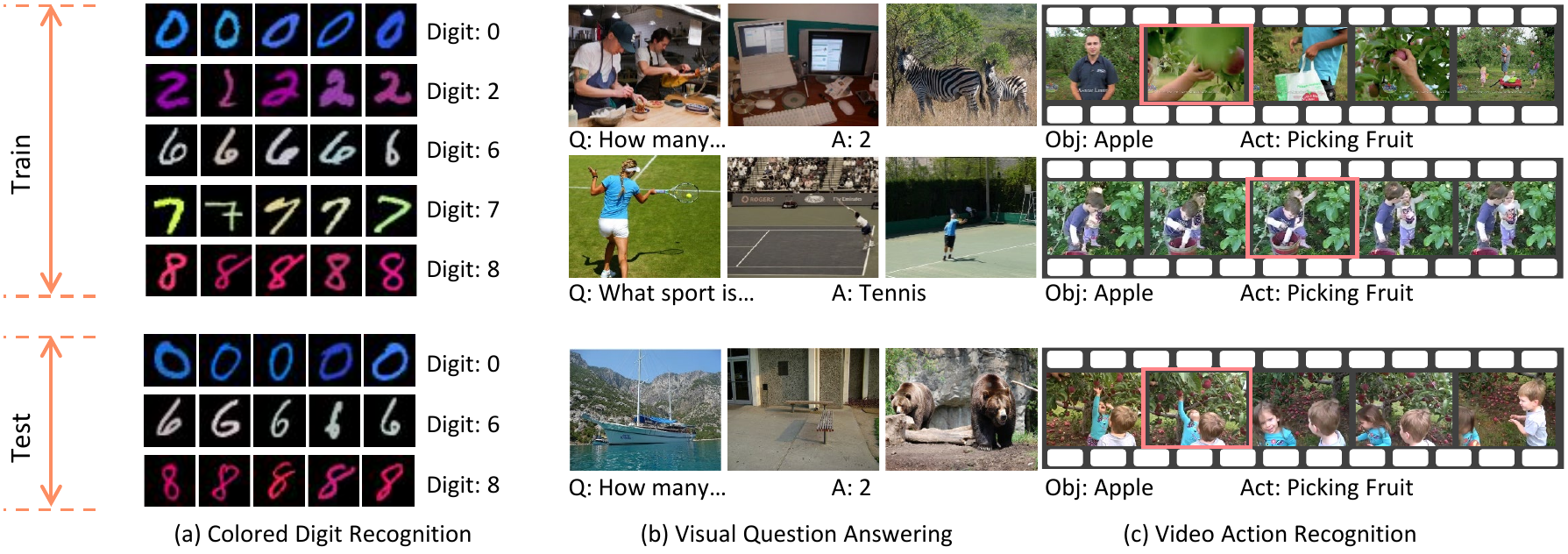}
  \caption{Modality bias problem illustration over three multi-modal tasks. Both training and testing sets are independently and identically distributed, which drives the model decision during testing towards the modality bias in the training set. For (a), (b) and (c), the biased modality is color, language and object (frame), respectively.}\label{fig:teaser}
\end{figure}

The shortcut is due to the strong correlation between semantic labels and specific modalities for multi-modal classification. The language prior problem in Visual Question Answering (VQA)~\cite{vqacp, adavqa, film, adversarial-nips}, serves as one typical manifestation. It refers to blindly answering questions without performing visual reasoning over images, since there exists a spurious connection between question types and answers (ref. \figref{fig:teaser}). In fact, this issue can be attributed to the modality bias influence, namely, one modality (language) dominates the class prediction than the other (vision). Inspired by this, we extend the problem to a broader scope and comprehensively study it from a larger perspective - the \emph{modality bias problem} in multi-modal learning. 

The modality bias problem is noticeably common during multi-modal learning. As illustrated in the two examples from \figref{fig:teaser}(a) and (c): for the colored digit recognition, the \emph{color} modality overwhelms the \emph{shape}~\cite{mfe} when predicting digits; And the modality \emph{motion} is restrained by the \emph{frame} for video action recognition. As a matter of fact, it poses several drawbacks to the model learning: the robustness and explainability of the existing methods are largely limited; generalization across datasets becomes impossible, to name a few. Nevertheless, this problem has been largely unaddressed thus far to the best of our knowledge. Towards this end, this work tentatively highlights the modality bias problem for the multimedia research community. 

Evaluating such a problem is rather difficult, as a simple model can achieve satisfactory performance over current benchmarks due to the Independently and Identically Distributed (I.I.D) property of training and testing sets. Therefore, the comparison among methods is oblivious towards the alleviation of the modality bias problem. Thanks to the recent progress on the Out-of-Distribution (OoD) generalization, Agrawal \etal~\cite{vqacp} proposed to re-split the VQA datasets to build their associated OoD counterparts. In the new curated benchmarks, the training and testing sets have different prior distributions of answers for each question type. For instance, the answer \emph{2} is the most frequent one to \emph{how many} during training. By contrast, the answer \emph{1} becomes the dominating answer in the testing set. This phenomenon violates the I.I.D property which is viewed as standard by traditional machine learning algorithms. In this way, the modality bias is manually cut off since a VQA model trained on such datasets cannot leverage the shortcut between questions and answers to perform well. Following this paradigm, we further construct the OoD datasets for the colored digit recognition and video action recognition tasks. In particular, the label distribution with respect to the biased modality is made dissimilar between training and testing sets. As a result, when evaluating some strong baselines on these datasets, drastic performance degradation can be observed (ref. Section~\ref{preliminary}).

In addition to the OoD benchmark construction, we extend our previous work in~\cite{adavqa} to a novel and generic loss function, namely Multi-Modal De-Bias (dubbed as MMDB), from the viewpoint of feature space learning. To implement this, we transform the feature space from Euclidean to Cosine, whereinto the decision boundary is determined only by the angle between multi-modal fused feature vector and the final classification weight matrix. Specifically, an adaptive margin is introduced to achieve the goal that frequent and sparse classes take broader and tighter spaces, respectively. To evaluate the effectiveness of the proposed MMDB, we apply it to ten baselines in total across three typical multi-modal classification tasks. From the experimental results, our MMDB can enhance the baselines with significant performance gains, while introducing no inference overload.  

In summary, the contribution of this paper is four-fold:
\begin{itemize}
\item We systematically study the modality bias problem in the context of multi-modal classification. To the best of our knowledge, we are the first to present and investigate this problem for multi-modal learning from a comprehensive view.
\item To facilitate the evaluation of this problem, we  construct several Out-of-Distribution datasets for benchmarking purpose.
\item A novel multi-modal de-bias loss function method based on feature space learning is devised to reduce the modality bias problem. Notably, the proposed method is model agnostic which can be integrated into any existing approaches and demands zero-incremental inference time.
\item We apply this loss function to various baselines over three multi-modal classification tasks. When equipped with our method, promising performance improvements from baselines can be observed on the OoD datasets. As a side product, we achieve a new state-of-the-art on two publicly available VQA-CP benchmarks for the VQA task. The code has been released to facilitate further research along this line\footnote{https://github.com/guoyang9/AdaVQA.}.
\end{itemize}

Our prior work~\cite{adavqa} presents a de-bias loss function for tackling the language prior problem in VQA, and this paper extends it in the following aspects: 1) We formally define and comprehensively study the modality bias problem for multi-modal learning, while ~\cite{adavqa} focuses on the VQA task only. 2) We apply the de-bias loss function to two more tasks, \ie colored digit recognition and video action recognition. For the VQA task, another strong baseline - LXMERT~\cite{lxmert} is explored with the equipment of our loss function, and aiding us to achieve a new state-of-the-art on two VQA-CP benchmarks. 3) We provide an empirical explanation of the proposed de-bias loss function in Section~\ref{model}.

The rest of this article is organized as follows. In Section~\ref{related_work}, we briefly review the related literature. We then present the recognition and reduction of the modality bias problem in Section~\ref{preliminary} and \ref{model}, respectively. In the next, the experimental settings are detailed in Section~\ref{setting}, followed by the results over the three tasks in Section~\ref{results}. We summarize this paper and discuss the possible future work in Section~\ref{conclusion}.

%% file: segments/related_work.tex
\section{Related Work}\label{related_work}
\subsection{Bias Identification and Mitigation}
The bias problem has long been recognized as an issue of concern in AI algorithms~\cite{age_bias, texture_bias1, 3d, domain}. Though effectively leveraging the bias can achieve acceptable results, the methods become less reliable and less robust to generalize over diverse datasets. In the following, we exemplify this problem from both vision and language domains.

Existing studies often refer the bias in images or videos to certain unbalanced attributes~\cite{method_all}. For instance, researchers have discovered the age bias~\cite{age_bias} and texture bias~\cite{texture_bias1, texture_bias2} in image classification. And human faces exhibit strong racial bias~\cite{racial_bias} and gender bias~\cite{gender_bias}, both seriously damaging the face recognition accuracy. In addition, images in 3D faces are generally accompanied with different poses and lighting conditions~\cite{3d}. To tackle these challenges, studies have been devoted to transfer learning, domain adaptation~\cite{domain}, adversarial learning~\cite{adversarial} or utilizing external knowledge.

The most ubiquitous bias in the language is the semantic bias learned from large corpora. Language modeling, serves as the foundation of natural language processing, has extensively been proven to introduce discrimination in its embeddings~\cite{nlp_all1, nlp_all2}. These embeddings often involve unintended correlations and societal stereotypes (\eg connecting medical doctors more frequently to male than female~\cite{male_bias}). ~\citet{redbias} studied the bias from language modeling in the context of offensive contents detection. To address this problem, balancing dataset from the statistical view becomes much more popular. For instance, one can augment original data with external labeled data, oversampling or downsampling, sample weighting~\cite{external_data3}, and identity term swapping~\cite{external_data2}. ~\cite{external_data1} appends non-toxic samples containing identity terms from Wikipedia articles into the training data.

Orthogonal to the above mentioned methods, in this work, we explore the bias from the perspective of modalities. In fact, some modalities show strong correlation with labels than others. And learning on such data often results in severe over-fitting problem. To pinpoint the importance and influence of the issue, this paper, for the first time, comprehensively studies the modality bias problem in multi-modal learning.

\subsection{Language Prior Problem in VQA}
Considerable efforts have been devoted to the language prior problem in VQA, as most VQA models blindly answer questions without performing visual reasoning on images~\cite{vqacp, adversarial-nips, lpscore, unshuffle}. 
Current studies can be grouped into the following two categories. 

\textbf{Dataset Re-balancing.} Crowd-sourcing with human annotators makes the VQA datasets difficult to circumvent biases. Ever since the presentation of the first large-scale VQA dataset~\cite{vqa1}, the bias problem has impeded the development of more generally applicable methods. To amend this problem, ~\cite{vqacp, lpscore, loss-vqa} 
demonstrate that the bias still remains, which can potentially induce VQA models to learn language priors. In view of this, VQA-CP~\cite{vqacp} is later curated through data re-splitting. Consequently, the answer distribution of training and testing sets is distinct with respect to question types (\eg the most frequent answers in training and testing sets can be \emph{2} and \emph{1} for the question type \emph{how many}, respectively). The performance of many VQA models drops significantly on the VQA-CP datasets. More recent studies construct brand-new datasets following the answer distribution balancing rule to avoid the language prior problem~\cite{clevr}. 

\textbf{Model De-bias.} Balancing datasets is often time-consuming and labor-intensive, some methods thus make efforts to directly counter this problem, which mainly fall into two groups: single-branch and two-branch models. Specifically, single-branch methods are devised to enhance the visual feature learning in VQA~\cite{hint, critical}. For example, HINT~\cite{hint} and SCR~\cite{critical} align the region importance with the additionally collected human attention maps. VGQE~\cite{vgqe} considers the visual and textual modalities equally when encoding the question, where the question features include information from both modalities.  Differently, two-branch methods mostly introduce another question-only branch for deliberately capturing the language priors, followed by the question-image branch to restrain it. For example, Q-Adv~\cite{adversarial-nips} trains the above two models in an adversarial way, which minimizes the loss of the question-image model while maximizing the question-only one. More recent fusion-based methods~\cite{lmh, rubi, counterfactual} employ the late fusion strategy to combine the two predictors and guide the model to pay more attention on these answers which cannot be correctly addressed by the question-only branch.

\subsection{Out-of-Distribution Generalization}
The I.I.D (Independently and Identically Distributed) property is deemed as \emph{de facto} in traditional machine learning algorithms. However, some studies point out that the robustness and generalization are greatly limited by the distribution shifts~\cite{ood1, ood2}. To deal with this, Out-of-Distribution evaluation has been evoked, wherein the testing data come from a distribution different with that of the training data. For instance, Benjamin \etal~\cite{imgnetv2} built the ImageNetV2 benchmark to maintain the naturally occurring distribution shift. Previous strong baselines exhibit a dramatic performance drop on this dataset~\cite{imgnetv2method}. In addition, some methods approach this new challenge with domain-invariant learning~\cite{ooddomain}, feature decomposition~\cite{oodfeature} and pre-training~\cite{oodpretrain}.

The OoD property offers a desirable criterion to test a model's generalization capability in real-world data. In view of this, we for the first time, curated two Out-of-Distribution dataset versions for colored digit recognition and video action recognition. These benchmarks are essential to diagnose the modality bias problem and well support its evaluation.   

%% file: segments/definition.tex
\section{Modality Bias Problem Recognition} \label{preliminary}
\begin{table}
  \centering
  \caption{Accuracy comparison from In-Domain and OoD evaluations for the Visual Question Answering task.}\label{tab:vqa_base}
  \begin{subtable}[c]{.9\textwidth}
  \centering
  \caption{Accuracy comparison on VQA v2 dataset and its OoD version VQA-CP v2.}
  \begin{tabular}{r|cccc|cccc}
    \toprule
    \multirow{2}{*}{Method}     & \multicolumn{4}{c|}{VQA v2 Val (In-Domain)}      & \multicolumn{4}{c}{VQA-CP v2 Test (OoD)} \\
                                  \cmidrule(lr){2-5}                                \cmidrule(lr){6-9}
                                & Y/N       & Num.      & Other     & All           & Y/N       & Num.      & Other     & All                  \\
    \midrule
    Counter~\cite{counter}      & 81.63     & 47.12     & 56.54     & 64.73         & 41.01     & 12.98     & 42.69     & 37.67                 \\
    UpDown~\cite{updown}        & 79.87     & 41.73     & 52.29     & 61.27         & 49.78     & 14.07     & 43.42     & 40.79                 \\
    LXMERT~\cite{lxmert}        & 88.34     & 56.64     & 65.78     & 73.06         & 46.70     & 27.14     & 61.20     & 51.78                 \\
    \bottomrule 
  \end{tabular}
  \end{subtable}
  \begin{subtable}[c]{.9\textwidth}
  \centering
  \caption{Accuracy comparison on VQA v1 dataset and its OoD version VQA-CP v1.}
  \begin{tabular}{r|cccc|cccc}
    \toprule
    \multirow{2}{*}{Method}     & \multicolumn{4}{c|}{VQA v1 Val (In-Domain)}      & \multicolumn{4}{c}{VQA-CP v1 Test (OoD)} \\
                                  \cmidrule(lr){2-5}                                \cmidrule(lr){6-9}
                                & Y/N       & Num.      & Other     & All           & Y/N       & Num.      & Other     & All                  \\
    \midrule
    Counter~\cite{counter}      & 84.39     & 42.12     & 56.29     & 65.03         & 39.12     & 13.09     & 42.35     & 36.11                 \\
    UpDown~\cite{updown}        & 82.58     & 37.81     & 51.59     & 61.46         & 43.76     & 12.49     & 42.57     & 38.02                 \\
    LXMERT~\cite{lxmert}        & 79.79     & 40.59     & 62.00     & 65.97         & 54.08     & 25.05     & 62.72     & 52.82                 \\
    \bottomrule
  \end{tabular}
  \end{subtable}
\end{table}

A good multi-modal model is expected to do prediction using informative features from all modalities. Existing methods have pushed the boundaries of various multi-modal benchmarks. Nevertheless, the improved performance is actually somewhat misleading, as both training and testing sets follow the I.I.D property. In this way, a model fitted on the training set may take a shortcut to perform well on its counterpart testing set. One undesirable shortcut recognized by this paper is the bias inherent in modalities, which refers to making prediction based on the correlation between certain factors from one modality and the labels. Take the colored digit recognition task as an example, if all the \emph{0} digits are colored  \emph{blue}, it is effortless for the current strong deep learning models to bias on this \emph{color} modality while ignoring the discriminative \emph{shape} modality.

In the following, we illustrate this problem from two aspects: performance degradation on OoD datasets and prediction towards modality bias. 

\subsection{Performance Degradation on OoD Datasets} The past few years have witnessed an increasing interest in OoD generalization~\cite{ood1, ood2}. It offers a strong test bed for evaluating the generalization capability of existing biased methods. As a matter of fact, the label distribution between training and testing sets is distinct from each other. Agrawal \etal~\cite{vqacp} curated the OoD version of traditional VQA datasets~\cite{vqa1, vqa2}. In particular, the answer distributions of each question type are significantly different between the training and testing sets. We re-implemented three baselines, including two well-studied methods (\ie Counter~\cite{counter} and UpDn~\cite{updown}) and a recently developed BERT-based one (LXMERT~\cite{lxmert}), and tested their performance on both in-domain and OoD versions of two VQA datasets. The results in \tabref{tab:vqa_base} demonstrate that the performance of all these methods drops drastically on the OoD datasets (see the \emph{All} category). For instance, the performance degradation of Counter is almost half on the two VQA-CP datasets with respect to the \emph{All} category. This phenomenon is mainly attributed to the blind model learning, \ie answering questions without the visual information. Pertaining to the in-domain dataset, the answer distribution under question types are consistent for both training and testing sets, \eg \emph{2} is the most frequent answer for \emph{how many}. When it comes to OoD , \emph{how many} questions may frequently correspond to \emph{1} in the testing set. Therefore, models leveraging such shortcuts suffer on this dataset.

As for the colored MNIST dataset, we constructed its OoD version based on the rule that the colors for each digit are made distinct in the training and testing sets. We leveraged three methods to demonstrate the results of this problem in \tabref{tab:mnist_base}. Similar observations can be seen from this task as the correlation between the \emph{color} modality and labels is cut off. As a result, models cannot generalize well on this dataset due to the modality bias learning. 
\begin{table}[htbp]
  \centering
  \caption{Accuracy comparison from In-Domain and OoD evaluations on the colored MNIST dataset.}\label{tab:mnist_base}
  \scalebox{1.0}{
  \begin{tabular}{r|cc|cc}
    \toprule
    \multirow{2}{*}{Method} & \multicolumn{2}{c|}{In-Domain}       & \multicolumn{2}{c}{OoD}   \\
                              \cmidrule(lr){2-3}                    \cmidrule(lr){4-5}
                            & ACC               & Loss              & ACC               & Loss                  \\
    \midrule
    MLPs                    & $99.18 \pm .17$   & $0.05 \pm .02$    & $55.55 \pm 1.36$  & $1.31 \pm .03$        \\
    LeNet~\cite{lenet}      & $99.36 \pm .15$   & $0.03 \pm .01$    & $57.39 \pm 11.22$ & $1.38 \pm .12$        \\
    ResNet18~\cite{resnet}  & $99.02 \pm .53$   & $0.03 \pm .01$    & $40.19 \pm 12.71$ & $1.57 \pm .16$        \\
    \bottomrule
  \end{tabular}
  }
\end{table}

In addition, we also constructed the OoD dataset for Kinetics-400 Kinetics-700, two widely-exploited benchmarks for video action recognition. Specifically, we made the action distribution with respect to the detected object different in the re-constructed dataset. For example, the most frequent action for object \emph{apple} is \emph{picking fruit} in the training set while it is other actions in the testing set. In this way, the bias from the \emph{frame} modality is manually removed to some extent. With this operation, we evaluated the I3D network~\cite{i3d} under two backbones and show the results in \tabref{tab:action_base}. The model's performance also drops though the degradation is not as severe as in the above two tasks. One reason might be that the actions are relatively balanced with the most dominant objects \emph{human}.
\begin{table}[htbp]
  \centering
  \caption{Performance comparison from In-Domain and OoD evaluations  on the Kinetics-400 dataset.}\label{tab:action_base}
  \scalebox{1.0}{
  \begin{tabular}{r|cc|cc}
    \toprule
    \multirow{2}{*}{Method} & \multicolumn{2}{c|}{In-Domain}        & \multicolumn{2}{c}{OoD}   \\
                              \cmidrule(lr){2-3}                    \cmidrule(lr){4-5}
                            & Precision@1       & Precision@5       & Precision@1       & Precision@5           \\
    \midrule
    I3D-ResNet50            & 55.92             & 80.53             & 52.20             & 80.47                 \\
    I3D-ResNet101           & 57.70             & 81.59             & 53.64             & 81.54                 \\
    \bottomrule
  \end{tabular}
  }
\end{table}

\subsection{Prediction towards Modality Bias}
\begin{figure}
  \centering
  \includegraphics[width=0.9\linewidth]{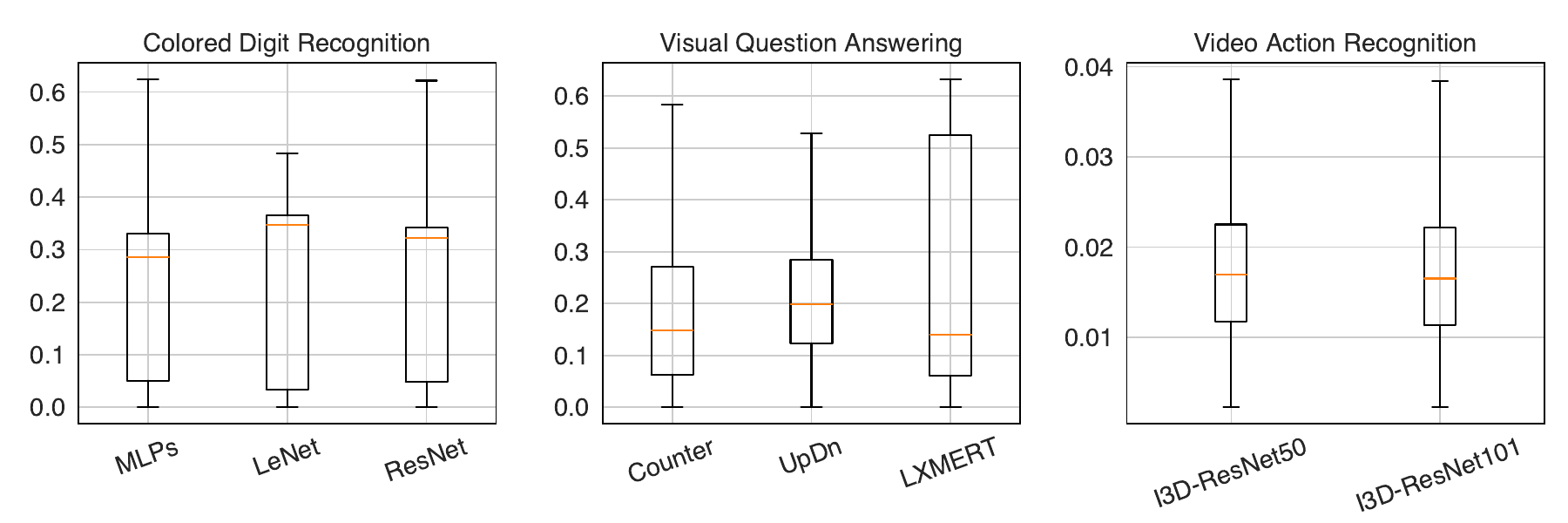}
  \caption{Jensen–Shannon Divergence values computed from two distributions: the label distribution with respect to the biased modality (\ie \emph{color} for colored digit recognition, \emph{question type} for VQA and \emph{object} for video action recognition) and the model outputted scores of incorrect instances.}\label{fig:jsd}
\end{figure}

To further understand how the models predict towards the modality bias, we computed the Jensen–Shannon Divergence (JSD) values of the biased label distribution and the model output of wrongly predicted instances. The results are illustrated in \figref{fig:jsd}. Ideally, without any biases, the predicted wrong labels from models should be more diverse or roughly follow the uniform distribution over all classes instead of being proportional to the label distribution conditioned on the biased modality in the training set. However, as shown in this figure, we can observe that most JSD values are below 0.5, implying that the two distributions are very similar. This indicates that models tend to provide labels according to the patterns observed between the biased modality and labels in the training set, rather than performing reasoning for the current instance. For example, a large portion of images with \emph{blue} digits are mis-predicted to \emph{0} for the colored digit recognition task, which also corresponds to the most frequent digits with the \emph{blue} color in the training set. 

%% file: segments/method.tex
\section{Modality Bias Reduction}\label{model}
A typical multi-modal classification model can be abstracted into three consecutive stages: a) multi-modal input representation; b) multi-modal fusion and c) classifier. The first stage takes as inputs the raw multi-modal data and outputs the embedded features. Thereafter, the features are fused with delicately designed manners yet not limited to simple concatenation~\cite{concate} and addition~\cite{addition}. Finally, a cross entropy loss function is employed to map the fused features to one or a few classes. It is worth noting that the modality bias problem can be triggered by any of the above three stages. In this work, from a generic view, we aim to tackle this problem based on classifier balancing and design a novel loss function. To this end, our objective becomes the frequent and sparse classes respectively taking broader and tighter feature spans in the final feature space, respectively.   
\subsection{Formulation Background}
Following the prevalent formulation, we consider the multi-modal classification as a multi-class single-/multi-label classification problem. That is, for input data with multiple modalities $M=\{M_1, M_2, ... M_n\}$, the objective function is given by:
\begin{equation} \label{equ:definition}
    \hat{y} = \mathop{\arg \max}_{y \in \Omega}  p(y|M_1, M_2, ... M_n; \Theta),
\end{equation}
where $\Omega$ and $\Theta$ denote the available class set and the model parameters, respectively. 

\textbf{SoftMax.} The most popular cross entropy loss function (we tag it as SoftMax) is then formulated as\footnote{We utilize the individual loss as an illustration instead of the loss for total instances due to space limitation.}:
\begin{equation} \label{equ:softmax}
\begin{aligned}
    L_{softmax} &= \sum_{i=1}^{|\Omega|} - y_i \log p_i \\
        &= \sum_{i=1}^{|\Omega|} - y_i \log \frac{\exp (\mathbf{W}_i^T \mathbf{x})}{\sum_{j=1}^{|\Omega|} \exp (\mathbf{W}_j^T \mathbf{x})},
\end{aligned}
\end{equation}
where $\mathbf{W}$ and $\mathbf{x}$ denote the weight matrix and feature vector directly adjacent to the class prediction, respectively. Note that for single-label classification, there is only one label $y_i$ equals 1 while others in the label vector $\bf{y}$ are kept as 0. When it comes to multi-label scenario, the involved ground-truth can be smoothed within the range of $(0, 1]$. Besides, we remove the bias vector for simplicity as we  found it contributes little to the final model performance. 

\textbf{NSL.} Recently, some studies have been dedicated to challenging the domination of traditional SoftMax loss function for classification tasks~\cite{arcface, cosface}. Among these efforts, switching from Euclidean space to Cosine space has been proven to be an intriguing fashion, which employs $L2$ normalization on both the final features as well as the weight vectors. By removing radial variations, it relieves the need for joint supervision of the norm and angle from the SoftMax loss. To approach this idea, we firstly leverage the $L2$ normalization on weight vector $\mathbf{W}_i$ and feature vector $\mathbf{x}$~\cite{alberto}. It is leveraged to ensure the posterior probability to be determined by the angle $\theta_i$ between $\mathbf{W}_i$ and $\mathbf{x}$, \ie $||\mathbf{W}_i||_2=1$ and $||\mathbf{x}||_2=1$ given the condition that $\mathbf{W}_i^T \mathbf{x} = ||\mathbf{W}_i|| ||\mathbf{x}|| cos \theta_i$. Accordingly, the feature space is converted from the Euclidean space to the Cosine one. We then provide the modified normalized SoftMax loss (NSL)~\cite{cosface} as follows,
\begin{equation} \label{equ:ns}
    L_{nsl} = \sum_{i=1}^{|\Omega|} - y_i \log \frac{\exp{(s \times \cos{\theta_i})}}{\sum_{j=1}^{|\Omega|} \exp{(s \times \cos{\theta_j})}},
\end{equation}
where $s$ is a scale factor for more stable computation. 

\textbf{LMCL.} In order to achieve a more discriminative classification boundary, LMCL~\cite{cosface} introduces a fixed cosine margin to NSL,  
\begin{equation} \label{equ:lmlc}
    L_{lmcl} = \sum_{i=1}^{|\Omega|} - y_i \log \frac{\exp{s ( \cos{\theta_{i}} - m)}}{\sum_{j \neq i} \exp{s \times \cos{\theta_j}} + \exp{s (\cos{\theta_{i}} - m)}},
\end{equation}
where $m$ implies the fixed cosine margin. Compared to the Euclidean space, the Cosine space is relatively easy to manipulate, as the margin range reduces from $(-\infty, +\infty)$ to $[-1, 1]$.

Regarding the implementation, we found that applying a fixed cosine margin cannot obtain satisfactory results. The key reason is that label distribution is highly skewed given the biased modality, resulting in the incapability of learning a sufficient representation with a fixed margin in the Cosine space. In the next subsection, we will introduce a more sophisticated adapted margin cosine loss to overcome this issue. 
\subsection{Proposed Method}
The results in our experiments (see Sect.~\ref{exp:ablation}) explicitly demonstrate that a fixed cosine margin yields limited improvements or even degrades the model performance. Based upon this observation, we argue that an adapted cosine margin is more favorable for tackling the bias problem in multi-modal classification. In view of this, a new loss function named Multi-Modal De-Bias (MMDB) is defined as:
\begin{equation} \label{equ:mmdb}
\left\{ 
\begin{aligned}
    & L_{MMDB} = \sum_{i=1}^{|\Omega|} - y_i \log \frac{\exp{s (\cos{\theta_{i}} - m_i)}}{\sum_{j=1}^{||\Omega||} \exp{s (\cos{\theta_j} - m_j)}},
      \\
    & m_i = 1 - \bar{m}_i, \\
    & \bar{m}_i = \frac{n_i^k + \epsilon}{\sum_{j=1}^{|\Omega|}n_j^k + \epsilon},
\end{aligned}
\right.
\end{equation}
where $m_i$ is the adapted margin for label $i$ which is estimated solely on the biased modality, $n_i^k$ denotes the number of label $i$ under the biased modality $b_k$ (\eg \emph{color} for \emph{colored digit recognition}) in the training set, $\epsilon = 1e-6$ is a hyper-parameter for avoiding computational overflow. The underlying intuition is that, for the current given biased modality, the frequent classes span broader in the Cosine space (smaller margin), while sparse classes span tighter (larger margin). In other words, frequent classes imply more training samples, which require a broader feature space to sufficiently cover these classes. In contrast, a tighter feature space is acceptable for sparse classes as the number of training samples is much smaller. This setting enables the models to place a better margin in the Cosine feature space. 

\begin{figure}
  \centering
  \includegraphics[width=0.9\linewidth]{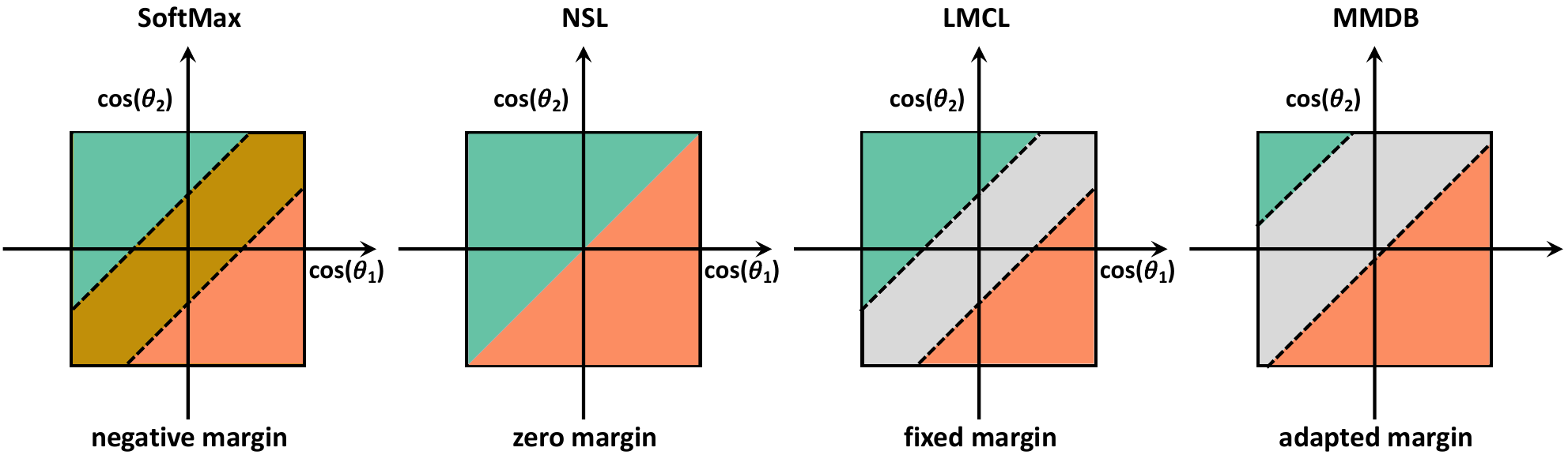}
  \caption{The visual comparison of decision boundary from different loss functions.}\label{fig:model}
\end{figure}

\textbf{Partial derivatives.} We further provide the partial derivatives of the weight vector $\mathbf{W}_i$ and feature vector $\mathbf{x}$ from our loss function. Let $p_i = s (\cos{\theta_{i}} - m_i) = s(\frac{\mathbf{W}_i^T}{||\mathbf{W}_i||_2} \cdot \frac{\mathbf{x}}{||\mathbf{x}||_2} - m_i)$, and $\hat{p}_i = \frac{\exp{p_i}}{\sum_{j=1}^{|\Omega|}\exp{p_j}}$. The partial derivatives are obtained via,
\begin{equation}
    \frac{\partial L_{MMDB}}{\partial \mathbf{x}} =  \sum_{i=1}^{|\Omega|} (\sum_{j=1}^{|\Omega|} y_j \times \hat{p}_i - y_i) \times s \times  \frac{||\mathbf{x}||_2 \mathbf{I} - \mathbf{x}\mathbf{x}^T}{||\mathbf{x}||_2^3} \frac{\mathbf{W}_i}{||\mathbf{W}_i||_2},
\end{equation}
and
\begin{equation}
    \frac{\partial L_{MMDB}}{\partial \mathbf{W}_i} =  \sum_{i=1}^{|\Omega|} (\sum_{j=1}^{|\Omega|} y_j \times \hat{p}_i - y_i) \times s \times  \frac{||\mathbf{W}_i||_2 \mathbf{I} - \mathbf{W}_i \mathbf{W}_i^T}{||\mathbf{W}_i||_2^3} \frac{\mathbf{x}}{||\mathbf{x}||_2}.
\end{equation}

\textbf{Lower bound for s.} The scale factor $s$ is critical for the final feature learning. A too small $s$ leads to an insufficient convergence as it limits the feature space span (as we found in our experiments that the loss goes `nan' with a small $s$). In view of this, a lower bound for $s$  should be prescribed. Without loss of generality, let $P_{i}$ denote the expected minimum of the class $i$, the lower bound is defined as,  
\begin{equation} \label{equ:scale}
     s \geq \frac{\ln(\nicefrac{1}{P_{i}} - 1)}{m_{i} + \nicefrac{\sum_{j \neq i} m_j}{(|\Omega| - 1}) - 2}.
\end{equation}
The detailed proof is provided in Appendix~\ref{proof}.

\subsection{Application over Specific Tasks}
We apply our MMDB loss function to three tasks according to their distinctive characteristics.

\textbf{Colored Digit Recognition.} As discussed in Section~\ref{preliminary}, compared to the \emph{shape} modality, the \emph{color} serves as the key biased factor in this task. For instance, most \emph{0} digits correspond to the \emph{blue} color (ref. \figref{fig:teaser}). In view of this, we compute the margin $\bar{m}_i$ with the constraint of colors, namely, $n_i^k$ is the number of digit $i$ under the color $k$.

\textbf{Visual Question Answering.} Some studies have been conducted on the language prior problem in VQA~\cite{vqacp, adversarial-nips, adavqa}. The language shortcut is deemed as the bias factor, where its expression is the strong link between question type (the first few words in a question) and the textual answer. Motivated by this observation, $n_i^k$ in \eqnref{equ:mmdb} can be simply estimated by the number of answer $i$ under the question type $k$. As a result, for the given question and its corresponding question type, the frequent answers span broader in the Cosine space (smaller margin) while sparse answers learn tighter feature space (larger margin). 

\textbf{Video Action Recognition.} Though actions in videos are expected to be recognized with the temporal information, nevertheless, we find that some of them are easy to be classified with the spatial modality, or more specifically, the \emph{objects} in static frames. To this end, we leverage the number of action $i$ under the detected object $k$ to compute the $n_i^k$, which is expected to alleviate the modality bias problem within this task.

\begin{figure}
  \centering
  \includegraphics[width=1.0\linewidth]{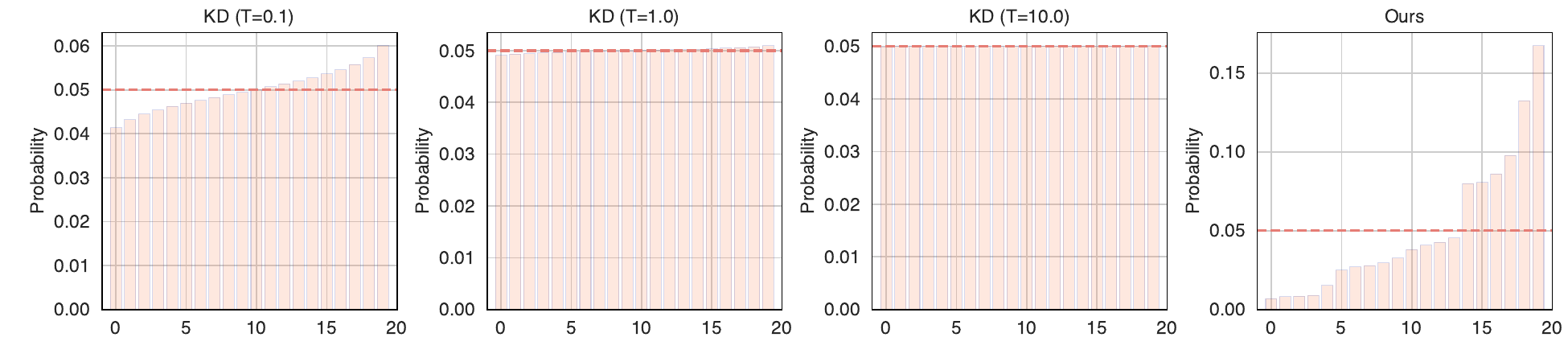}
  \caption{Probability comparison from 20 classes between KD and our method where the probabilities are arranged from small to large. The left three ones respectively denote the KD with temperature 0.1, 1.0 (\ie SoftMax) and 10.0. And the right most illustrates the probability yields from our method. The averaged probability from the 20 classes are outlined in red for all cases.}\label{fig:softmax}
\end{figure}

\subsection{Comparison with Different Loss Functions}
We consider the binary-class scenario for intuitively illustrating the decision boundary from different loss functions. As can be seen from \figref{fig:model}, the decision boundary of the plain SoftMax one can be negative, which is enlarged to be equal to zero of the NSL loss~\cite{cosface}. The LMLC~\cite{cosface} defines a fixed margin for different classes, yet is not suitable in our case for overcoming the modality bias problem. Regarding our MMDB, a sparser class (green one) is mapped to a smaller feature space while a more frequent class (orange one) engages with a larger feature space.
\subsection{An Empirical Explanation}
To understand why the proposed loss function works, recall that $\mathbf{W}_i^T \mathbf{x} = ||\mathbf{W}_i|| \times ||\mathbf{x}|| \times \cos{\theta_i} = s \times \cos{\theta_i}$, we firstly unfold $L_{MMDB}$ with:
\begin{equation}
\begin{aligned}
    L_{MMDB}    &= \sum_{i=1}^{|\Omega|} - y_i \log \frac{\exp{s (\cos{\theta_{i}} - m_i)}}{\sum_{j=1}^{||\Omega||} \exp{s (\cos{\theta_j} - m_j)}} \\
                &= \sum_{i=1}^{|\Omega|} - y_i \log \frac{\exp{(\mathbf{W}_i^T \mathbf{x} - s \times m_i)}}{\sum_{j=1}^{||\Omega||} \exp{(\mathbf{W}_j^T \mathbf{x} - s \times m_j)}} \\
                &= \sum_{i=1}^{|\Omega|} - y_i \log \frac{\nicefrac{\exp{(\mathbf{W}_i^T \mathbf{x}})}{\exp{(s \times m_i)}}}{\sum_{j=1}^{||\Omega||} \nicefrac{\exp{(\mathbf{W}_j^T \mathbf{x})}}{\exp{(s \times m_j)}}}.
\end{aligned}
\end{equation}
We then rewrite,
\begin{equation}
    T_i = \exp{(s \times m_i)},
\end{equation}
where we name $T_i$ as the temperature for class $i$, which produces,
\begin{equation}
    L_{MMDB} = \sum_{i=1}^{|\Omega|} - y_i \log \frac{\nicefrac{\exp{(\mathbf{W}_i^T \mathbf{x})}}{T_i}}{\sum_{j=1}^{||\Omega||} \nicefrac{\exp{(\mathbf{W}_j^T \mathbf{x})}}{T_j}}.
\end{equation}
Note that the logits are softened in the well-studied knowledge distillation (KD) domain~\cite{kd, kd-hinton} through,
\begin{equation}
    q_i = \frac{\exp (\nicefrac{\mathbf{W}_i^T \mathbf{x}}{T})} {\sum_{j=1}^{||\Omega||} \exp (\nicefrac{\mathbf{W}_j^T \mathbf{x}}{T})},
\end{equation}
where $T$ denotes the temperature during distilling and $q_i$ represents the probability for label $i$. There are two main differences between our method and KD: 1) we move the temperature outside the scope of the exponential function; and 2) the temperature is distinct for different labels. A visual comparison of the produced probabilities is demonstrated in \figref{fig:softmax}. We can observe that larger temperatures from KD leads to more balanced probabilities, which produces more ``informative'' labels as argued by~\cite{kd-hinton}. However, this may contradict the scenario studied in this paper, as the most ``informative'' classes to the current target are naturally the ones inducing the modality bias problem. In contrast, our method in \figref{fig:softmax} demonstrates a sharp probability distribution, which aids the reduction of this problem to some extent. 

%% file: segments/experimental_set.tex
\section{Experimental Setting}\label{setting}
We conducted extensive experiments on the aforementioned three multi-modal classification tasks, to validate the effectiveness of the proposed method. In particular, the experiments are mainly performed to answer the following research questions:
\begin{itemize}
  \item \textbf{RQ1}: Can the proposed multi-modal de-bias loss function method overcome the modality bias problem?
  \item \textbf{RQ2}: How do fixed margins in \eqnref{equ:lmlc} and the scale in \eqnref{equ:mmdb} affect the final model performance?
  \item \textbf{RQ3}: Why does the proposed method outperform the baselines?
\end{itemize}
To answer these questions, we firstly present the experimental setup for the three tasks. 
\subsection{Colored Digit Recognition}
\textbf{Dataset.} Li \etal~\cite{cmnist} firstly introduced the Colored MNIST dataset, where the color for the ten classes is made distinctive from each other. We observed that the training and testing sets still follow an I.I.D property, which is however, deficient to evaluate the modality bias problem. To this end, we proposed to perturb the color assignment for these two sets. In our experiments, we mainly tested the model performance on the newly curated OoD Colored MNIST dataset.  

\textbf{Evaluation Metric.} We adopted the standard accuracy metric for this experiment.

\textbf{Tested Baselines.} As the Colored MNIST dataset is relatively simple to address, therefore, a two-layer MLPs, LeNet~\cite{lenet} as well as a slightly cumbersome ResNet18~\cite{resnet} baselines are utilized. In addition, we also compared with two approaches working on the bias problem, \ie Repair~\cite{cmnist} and BiaSwap~\cite{biaswap}. And we ran each method for five times and reported the averaged accuracy. 
\subsection{Visual Question Answering}
\textbf{Datasets.}
We justified our proposed method on the two VQA-CP datasets: VQA-CP v2 and VQA-CP v1~\cite{vqacp}, which are widely accepted benchmarks for evaluating the models' capability to overcome the language prior problem. The VQA-CP v2 and VQA-CP v1 datasets consist of $\sim$122K images, $\sim$658K questions and $\sim$6.6M answers, and $\sim$122K images, $\sim$370K questions and $\sim$3.7M answers, respectively. Moreover, the answer distribution per question type is significantly different between training and testing sets (OoD property). For all the datasets, the answers are divided into three categories: \emph{Y/N}, \emph{Num.} and \emph{Other}.

\textbf{Evaluation Metric.}
We adopted the standard metric in VQA for evaluation~\cite{vqa1}. For each predicted answer $a$, the accuracy is computed as,
\begin{equation}\label{equ:metric}
  Acc = \text{min} (1, \frac{\#\text{humans that provide answer $a$}}{3}).
\end{equation}
Note that each question is answered by ten annotators, and this metric takes the human disagreement into consideration~\cite{vqa1, vqa2}.

\textbf{Tested Baselines.} Inherited from previous attention-based VQA models, Counter~\cite{counter} introduces a counting module to enable robust counting from object proposals. UpDn~\cite{updown} firstly leverages the pre-trained object detection frameworks to obtain salient object features for high-level reasoning. It then employs a simple attention network to focus on the most important objects which are highly related with the given question. In addition, the strong baseline LXMERT~\cite{lxmert} is built upon the Transformer encoders to learn the connections between vision and language. It is pre-trained with diverse pre-training tasks on several large-scale datasets of image and sentence pairs and achieves significant performance improvement on downstream tasks including VQA.

\subsection{Video Action Recognition}
\textbf{Dataset.} For this multi-modal task, we performed experiments on the widely exploited Kinetics-400 and Kinetics-700 datasets~\cite{kinetics}. In general, there are 400 and 700 action classes in these two datasets, respectively, with 400–1,150 clips for each action. And each clip is from a unique video. Each clip lasts around 10s and corresponds to a total number of 306,245 and 647,907 videos for the two datasets, respectively. 

\textbf{Evaluation Metric.} Followed previous video action recognition methods~\cite{action}, we adopted the clip-level accuracy, \eg Precision@1 and Precision@5 as the evaluation metrics. 

\textbf{Tested Baselines.} We employed our loss function over the I3D~\cite{i3d} network with two backbones - ResNet50 and ResNet101. As demonstrated in ~\cite{action}, this simple method performs on par with or even outperforms many recent methods that are claimed to be better. Due to the resource limitation, we reduced the number of frames per video and the batch size for both methods to 8 and 16, respectively, while kept other settings the same as the released code\footnote{https://github.com/IBM/action-recognition-pytorch.}. In addition, we also studied the effectiveness of our loss function on two recently developed strong baselines - TimeSformer~\cite{timesformer} and ViViT~\cite{vivit}.

It is worth noting that for all the baselines with respect to the above three tasks, we simply replaced the cross entropy loss with our MMDB and did NOT change any other settings, such as embedding size, learning rate, optimizer, and batch size. Therefore, our method will not introduce any incremental cost into those baseline models during the testing stage, since the inference of both our method and the baselines are identical.

%% file: segments/experimental_results.tex
\section{Experimental Results}\label{results}
\subsection{Overall Performance Comparison (RQ1)}
\subsubsection{Colored Digit Recognition.} 
\begin{figure}
  \centering
  \includegraphics[width=0.95\linewidth]{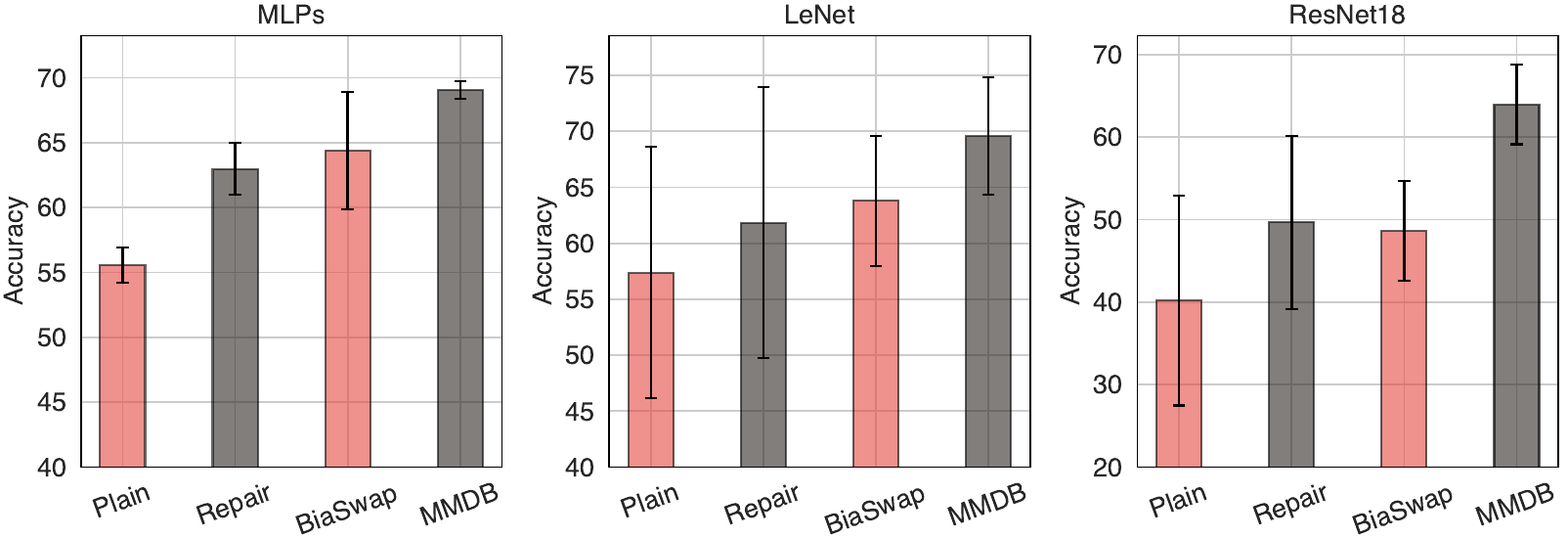}
  \caption{Performance comparison between our MMDB and three baselines on the colored digit recognition task. The error bars are also shown.}\label{fig:color-overall}
\end{figure}

We applied our method to three baselines: MLPs, LeNet and ResNet18 and report the results in \figref{fig:color-overall}. The observations are as follows:
\begin{itemize}
    \item All the three methods can achieve certain performance improvements over the plain baselines. Our method, \ie MMDB can outperform Repair~\cite{cmnist} and BiaSwap~\cite{biaswap} with significant gains on all the three baselines.
    \item Our method can significantly enhance the baselines with large performance margins. Specifically, the improved accuracy on MLPs, LeNet and ResNet18 are around 14\%, 12\% and 25\%, respectively. This demonstrates that the modality bias problem is reduced by our MMDB.
    \item Compared to the baselines, more stable error bar can be seen for MMDB. It shows that our method is more robust over multiple runs, which serves as another merit.  
\end{itemize}

\begin{table}[htbp]
  \centering
  \caption{Accuracy comparisons with respect to different answer categories over the VQA-CP v2 dataset. Regarding the method group, the top group denotes plain approaches, the middle group represents methods directly applied on the UpDn baseline, and the approaches from the last group are with our loss function. `$-$' and `$\dag$' denote the numbers are unavailable and our implementation, respectively. The best performance in current splits is highlighted in bold.}\label{tab:baseline-v2}
  \scalebox{1.0}{
  \begin{tabular}{r|cccc}
    \toprule
    \multirow{2}{*}{Method}             & \multicolumn{4}{c}{VQA-CP v2 test}        \\
                                        \cmidrule(lr){2-5}                     
                                        & Y/N       & Num.      & Other     & All   \\
    \midrule
    NMN~\cite{nmn}                      & 38.94     & 11.92     & 25.72     & 27.47 \\
    MCB~\cite{mcb}                      & 41.01     & 11.96     & 40.57     & 36.33 \\
    Counter\dag~\cite{counter}          & 41.01     & 12.98     & 42.69     & 37.67 \\
    UpDn~\cite{updown}                  & 42.27     & 11.93     & 46.05     & 39.74 \\
    UpDn\dag~\cite{updown}              & 49.78     & 14.07     & 43.42     & 40.79 \\
    
    LXMERT\dag~\cite{lxmert}            & 46.70     & 27.14     & 61.20     & 51.78 \\
    \midrule
    GVQA~\cite{vqacp}                   & 57.99     & 13.68     & 22.14     & 31.30 \\
    AdvReg~\cite{adversarial-nips}      & 65.49     & 15.48     & 35.48     & 41.17 \\
    Rubi~\cite{rubi}                    & 68.65     & 20.28     & 43.18     & 47.11 \\
    LMH~\cite{lmh}                      & -         & -         & -         & 52.05 \\
    LMH\dag~\cite{lmh}                  & 70.29     & 44.10     & 44.86     & 52.15 \\
    SCR~\cite{critical}                 & 72.36     & 10.93     & 48.02     & 49.45 \\
    VGQE~\cite{vgqe}                    & 66.35     & 27.08     & 46.77     & 50.11 \\
    CSS~\cite{counterfactual}           & 43.96     & 12.78     & 47.48     & 41.16 \\
    Decomp-LR~\cite{questiontype}       & 70.99     & 18.72     & 45.57     & 48.87 \\
    \midrule
    Counter+Ours                        & 61.00     & 53.22     & 43.17     & 49.90 \\
    UpDn+Ours                           & 72.47     & 53.81     & 45.58     & 54.67    \\
    LXMERT+Ours                         & \bf{91.37} & \bf{65.55} & \bf{62.61} & \bf{71.44}    \\
    \bottomrule
  \end{tabular}
  }
\end{table}
\begin{table}[htbp]
  \centering
  \caption{Accuracy comparisons with respect to different answer categories over the VQA-CP v1 dataset. `$\dag$' denotes our implementation. The best performance in current splits is highlighted in bold.}\label{tab:baseline-v1}
  \scalebox{1.0}{
  \begin{tabular}{r|cccc}
    \toprule
    \multirow{2}{*}{Method}             & \multicolumn{4}{c}{VQA-CP v1 test}        \\
                                        \cmidrule(lr){2-5}
                                        & Y/N       & Num.      & Other     & All   \\
    \midrule
    NMN~\cite{nmn}                      & 38.85     & 11.23     & 27.88     & 29.64 \\
    MCB~\cite{mcb}                      & 37.96     & 11.80     & 39.90     & 34.39 \\
    Counter\dag~\cite{counter}          & 40.93     & 12.87     & 42.72     & 37.08 \\
    UpDn\dag~\cite{updown}              & 43.76     & 12.49     & 42.57     & 38.02 \\
    GVQA~\cite{vqacp}                   & 64.72     & 11.87     & 24.86     & 39.23 \\
    AdvReg~\cite{adversarial-nips}      & 74.16     & 12.44     & 25.32     & 43.43 \\
    LMH\dag~\cite{lmh}                  & 76.61     & 29.05     & 43.38     & 54.76 \\
    LXMERT~\cite{lxmert}                & 54.08     & 25.05     & 62.72     & 52.82 \\
    \midrule
    Counter+Ours                        & 72.01     & 49.28     & 42.60     & 55.92 \\
    UpDn+Ours                           & 91.17     & 41.34     & 39.38     & 61.20\\
    LXMERT+Ours                         & \bf{92.67} & \bf{61.37} & \bf{64.04} & \bf{75.47} \\
    \bottomrule
  \end{tabular}
  }
\end{table}
\begin{table}[htbp]
  \centering
  \caption{Performance Comparison on the OoD version of the Kinetics-400 and Kinetics-700 datasets.}\label{tab:overall-video}
  \scalebox{1.0}{
  \begin{tabular}{r|r|cc|cc}
    \toprule
    \multirow{2}{*}{Dataset}    & \multirow{2}{*}{Method}       & \multicolumn{2}{c|}{Baseline}     & \multicolumn{2}{c}{MMDB}      \\
                                \cmidrule(lr){3-4}                    \cmidrule(lr){5-6}
                                                            &   & Precision@1   & Precision@5       & Precision@1   & Precision@5   \\
    \midrule
    \multirow{4}{*}{Kinetics-400} 
                                & I3D-ResNet50~\cite{i3d}       & 52.20             & 80.47             & 54.57             & 81.10 \\
                                & I3D-ResNet101~\cite{i3d}      & 53.64             & 81.54             & 55.66             & 81.82 \\
                                & TimeSformer~\cite{timesformer}& 72.01             & 90.75             & 73.65             & 91.36 \\
                                & ViViT~\cite{vivit}            & 75.24             & 93.25             & 75.39             & 93.50 \\
    \midrule
    \multirow{4}{*}{Kinetics-700} 
                                & I3D-ResNet50~\cite{i3d}       & 37.21             & 66.05             & 38.95             & 67.47 \\
                                & I3D-ResNet101~\cite{i3d}      & 30.52             & 59.20             & 34.60             & 62.25 \\
                                & TimeSformer~\cite{timesformer}& 61.33             & 83.90             & 65.65             & 86.28 \\
                                & ViViT~\cite{vivit}            & 72.00             & 91.41             & 74.83             & 92.01 \\
    \bottomrule
  \end{tabular}
}
\end{table}

\subsubsection{Visual Question Answering.} The experimental results on VQA-CP v2 and VQA-CP v1 are illustrated in \tabref{tab:baseline-v2} and \tabref{tab:baseline-v1}, respectively. The main observations from these two tables are listed below:
\begin{itemize}
    \item In general, the methods in the middle group of \tabref{tab:baseline-v2} (specially designed to overcome the language bias problem) often outperform previous strong baselines (\eg FiLM~\cite{film}). This result is intuitive as conventional approaches may introduce biases to model learning.
    \item Our method obtains the best results on the two benchmark datasets. With the help of the recent strong baseline LXMERT, it surprisingly achieves a new state-of-the-art on these two benchmarks.
    \item For all the three baselines, \ie Counter, UpDn and LXMERT, when equppied with our MMDB loss function, a drastic performance improvement (15\% on average) can be observed. For example, on the VQA-CP v2 dataset, LXMERT+Ours achieves an absolute performance gain of 19.66\% on the \emph{All} answer category; and on the VQA-CP v1 dataset, UpDn+Ours outperforms the baseline UpDn with 23.18\% on the \emph{All} category.
    \item Compared with other methods whose backbone model is also UpDn on the VQA-CP v2 dataset, our method (UpDn+Ours) still surpasses them by a large margin, especially for the three newly developed approaches VGQE, CSS and Decomp-LR.
\end{itemize}
\subsubsection{Video Action Recognition.} We tested the effectiveness of our method on the I3D network~\cite{i3d} with two backbones -- ResNet50 and ResNet101, TimeSformer~\cite{timesformer} and ViViT~\cite{vivit}. From the results in \tabref{tab:overall-video}, it can be seen that our method can boost the backbone with a significant improvement, especially for the Precision@1, the improvements are 2.27\% and 2.02\% for ResNet50 and ResNet101 on the Kinetics-400 dataset, respectively. 
\begin{table}
  \centering
  \caption{Effectiveness validation of the proposed loss function on three multi-modal learning tasks. For methods on Kenetics-400, we reported the Precision@1 metric.}\label{tab:fix-margin}
  \scalebox{0.95}{
  \begin{tabular}{cc|cccccccc}
    \toprule
    \multirow{2}{*}{Method} &\multirow{2}{*}{Margin}                 
                                            & \multicolumn{3}{c}{Colored MNIST} & \multicolumn{3}{c}{VQA-CP v2}     & \multicolumn{2}{c}{Kenetics-400}  \\
                                            \cmidrule(lr){3-5}                  \cmidrule(lr){6-8}                  \cmidrule(lr){9-10}
                        &                   & MLPs      & LeNet     & ResNet    & Counter   & UpDn      & LXMERT    & I3D-ResNet50  & I3D-ResNet101     \\
    \midrule
    Baseline            & -                 & 55.55     & 57.39     & 40.19     & 37.67     & 40.79     & 51.78     & 52.20         & 53.64             \\
    NSL                 & -                 & 56.48     & 55.93     & 48.70     & 49.08     & 40.97     & 58.06     & 53.95         & 55.45             \\
    \midrule
    \multirow{5}{*}{Fixed Margin}       
                        & 0.1               & 58.55     & 60.06     & 54.84     & 30.12     & 39.42     & 56.56     & 52.15         & 47.32             \\
                        & 0.3               & 19.92     & 58.18     & 52.73     & 27.70     & 37.75     & 55.05     & 46.54         & 46.78             \\
                        & 0.5               &  9.80     &  9.80     & 51.27     & 13.07     & 36.60     & 53.81     & 43.36         & 44.18             \\
                        & 0.7               &  9.80     &  9.80     & 53.80     & 11.41     & 35.83     & 53.19     & 42.71         & 43.31             \\
                        & 0.9               &  9.80     &  9.80     & 56.05     &  3.12     & 35.68     & 52.62     & 43.19         & 43.33             \\
    \midrule
    Adapted Margin      & \emph{adaptive}    & 68.05     & 69.59     & 64.00     & 49.90     & 54.67     & 71.42     & 54.57         & 55.66             \\ 
    \bottomrule
  \end{tabular}}
\end{table}

\begin{figure}
  \centering
  \includegraphics[width=0.95\linewidth]{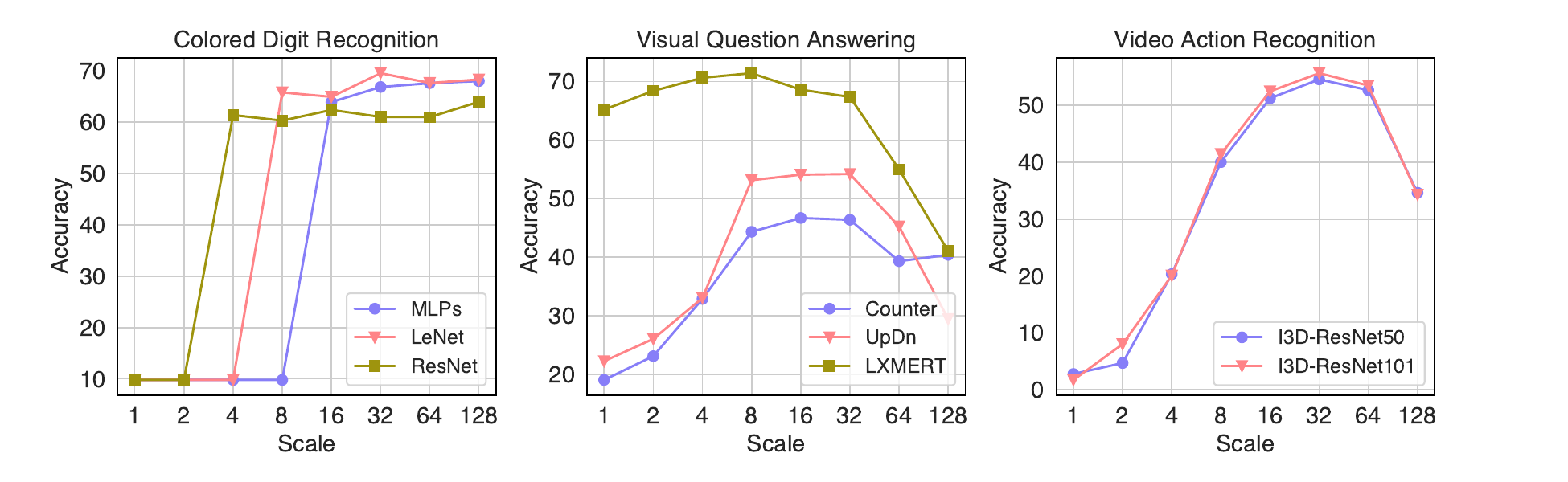}
  \caption{Performance change with respect to the scale $s$ in Eqn.~\ref{equ:mmdb}.}\label{fig:line}
\end{figure}

\subsection{Ablation Study (RQ2)}\label{exp:ablation} 
For a deeper understanding of our MMDB, we further provided detailed ablation studies over these three tasks. 

\textbf{Fixed margin results.} As mentioned in section~\ref{model}, models with a fixed margin perform unsatisfactorily when compared with our adaptive ones. The results are shown in \tabref{tab:fix-margin} and we have the following observations:
\begin{itemize}
\item When using the $L2$ normalization on both the weight vector and the feature vector, the \emph{de facto} NSL function, the results are inconsistent amongst different methods. For example, the Counter in VQA with NSL surpasses the baseline with 11.41\%, while LeNet in colored digit recognition with NSL even causes a little bit deterioration of the performance.
\item We also tested the fixed margin to yield effective feature discrimination, where the margins are tuned from 0.1 to 0.9 with a step size of 0.2. However, the results are not favorable (the improvement is limited), which validates the evidence that a fixed margin is not suitable for overcoming the modality bias problem. By contrast, when replacing the fixed margin with our adaptive one, the model can achieve significant performance improvement, which additionally proves the superiority of MMDB.
\end{itemize}
\begin{figure}
  \centering
  \begin{minipage}{0.47\linewidth}
  \centering
  \begin{minipage}{1.0\linewidth}
  \centering
  \includegraphics[width=1.0\linewidth]{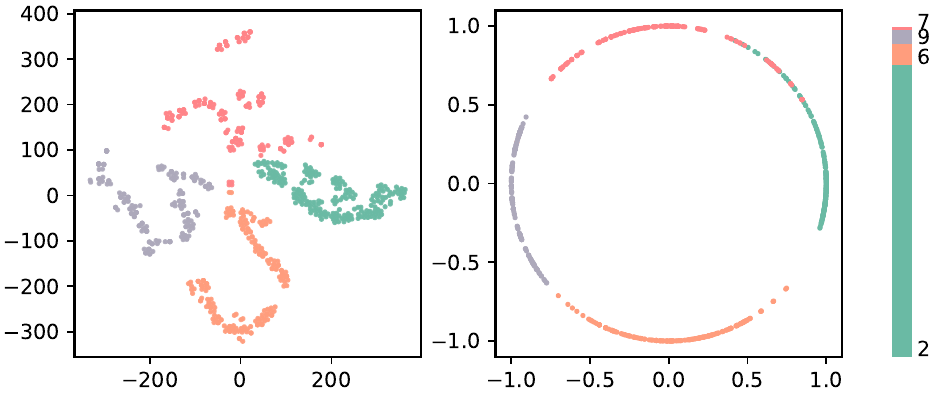}
  \end{minipage}
  \begin{minipage}{1.0\linewidth}
  \centering
  \includegraphics[width=1.0\linewidth]{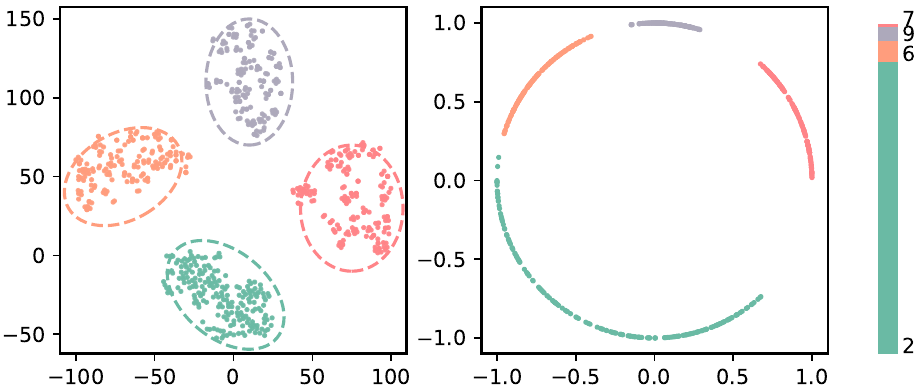}
  \end{minipage}
  \caption{Digit feature manifold embedding of \emph{red} color for the colored digit recognition task.}\label{fig:embed-color}
  \end{minipage}
  \begin{minipage}{0.48\linewidth}
  \centering
  \begin{minipage}{1.0\linewidth}
  \centering
  \includegraphics[width=1.0\linewidth]{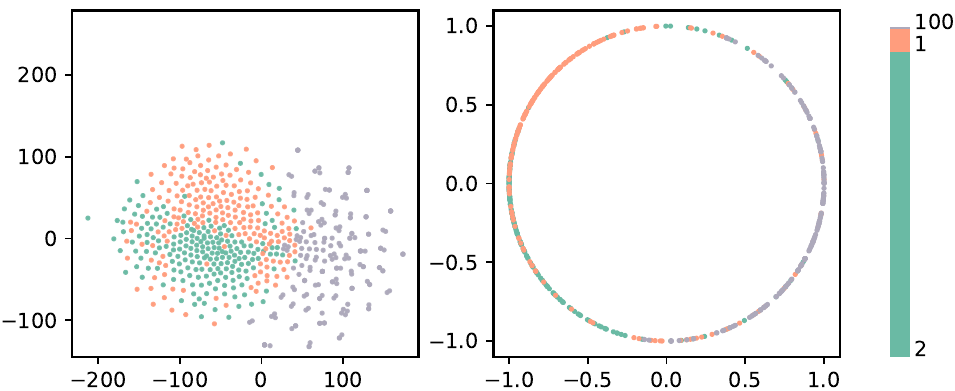}
  \end{minipage}
  \begin{minipage}{1.0\linewidth}
  \centering
  \includegraphics[width=1.0\linewidth]{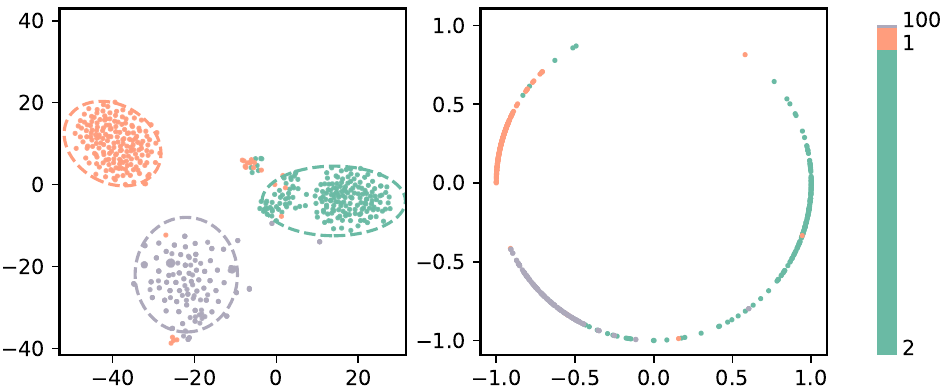}
  \end{minipage}
  \caption{Answer feature manifold embedding of \emph{how many} question type for the VQA task.}\label{fig:embed-vqa}
  \end{minipage}
 \label{fig:embed}
\end{figure}
\textbf{Scale influence.} We tuned the scale in \eqnref{equ:mmdb} from 1 to 128 with a step size of $2^n$ for all the eight methods. From the results in \figref{fig:line}, we found that a too small scale is insufficient for learning the feature space, resulting in unsatisfactory results. In addition, when the scale becomes large, the performance will saturate or even drop to some degrees. 
\subsection{Case Study (RQ3)}
In this subsection, we use case studies analyze the reasons why the proposed method works from the following two aspects: feature embedding separation and better attention maps.

\textbf{Feature manifold embedding.} Since the key motivation of the proposed method is to achieve the goal that frequent and sparse labels under certain biased modality span broader and tighter in the final feature space, respectively,  we therefore visualized the learned features and displayed the results on \figref{fig:embed-color} and \figref{fig:embed-vqa}. In particular, we leveraged two tasks - colored digit recognition and VQA for illustration. For the former task, the \emph{color} modality is restricted to \emph{red}, and the language modality in the later task is expressed through the \emph{how many} question type. For the two instances, the top two sub-figures are the feature embedding on the Euclidean and Cosine space from the baseline, while results from our method are summarized in the bottom two sub-figures. It is evident to us that 1) in the Euclidean space, the features are often irregular and even entangled with each other of baselines. Nevertheless, our method can separate these labels with clear boundaries. In addition, when it refers to the Cosine space, frequent labels (\eg digit \emph{2} for colored digit recognition and answer \emph{2} for VQA) span much broader than sparse ones, yet this property cannot be observed by the baseline.

\begin{figure}
  \centering
  \includegraphics[width=0.95\linewidth]{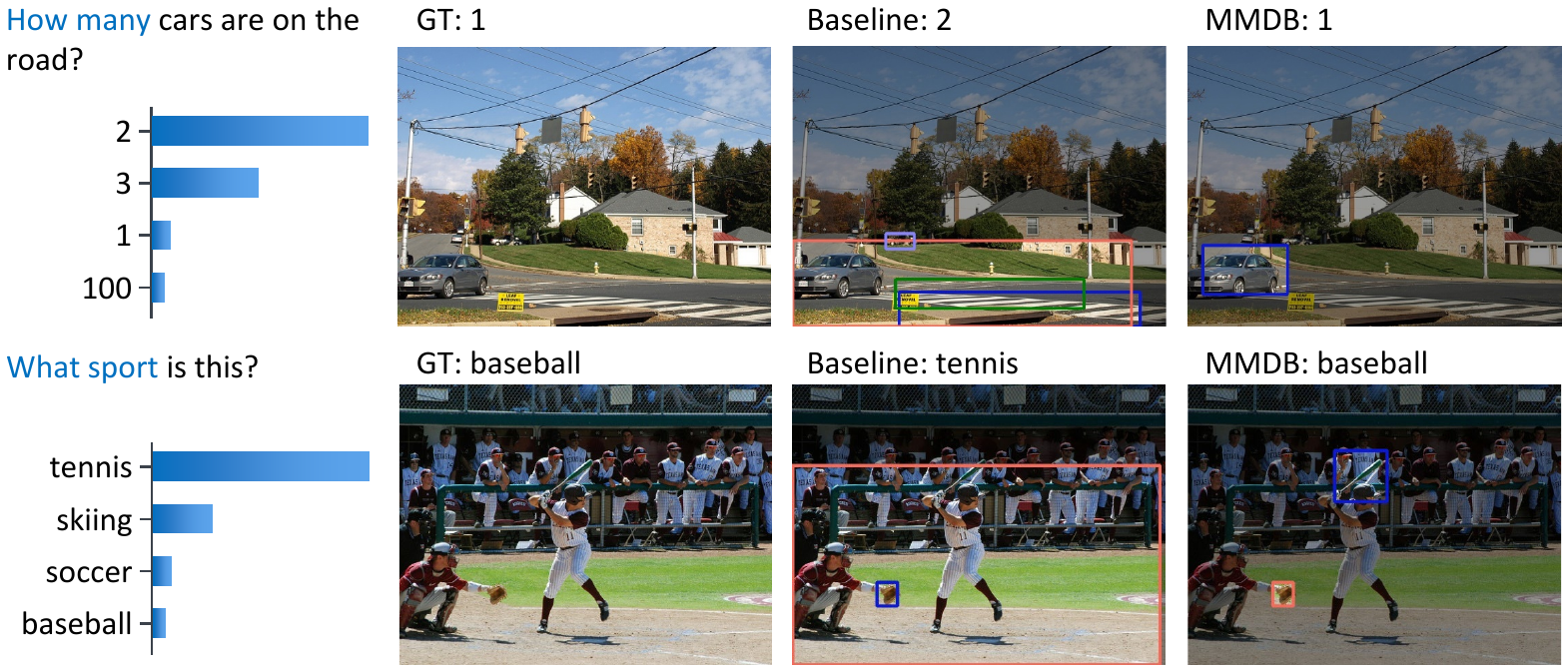}
  \caption{Visualization of the baselines method (UpDn) with and without our proposed loss function MMDB. The key answer distributions under the question's corresponding question type are illustrated on the leftmost column. The ground-truth is displayed on the second column, followed by the attention maps produced by the baseline and ours on the last two columns.}\label{fig:case-study}
\end{figure}

\textbf{Attention maps in VQA.} Finally, we showcased two successful cases of our method and illustrated them in \figref{fig:case-study}. Regarding the first case, as the answer \emph{2} takes large proportion under the question type \emph{how many} in the training set, the baseline model thus yields an answer \emph{2} to this question. In contrast, our MMDB corrects this mistake, producing a more reasonable attention map (focusing solely on one \emph{car}). As for the second case, the baseline model wrongly predicts the answer \emph{tennis} mainly attribute to \emph{tennis} being more frequent under the question type \emph{what sport} in the training set. Besides, too much attention has been paid to less relevant regions, which is another reason leading to the incorrect answer. On the contrary, our MMDB can guide the model to emphasize more on the target object - \emph{glove}, resulting in the correct answer \emph{baseball}.

%% file: segments/conclusion.tex
\section{Conclusion and Discussion}\label{conclusion}
In this work, we have systematically studied the modality bias problem in the context of multi-modal classification. To begin with, we construct several Our-of-Distribution datasets as test beds for its evaluation. Thereafter, a multi-modal de-bias loss is designed to discriminate the labels via properly characterizing the feature space. Concretely, for the given biased modality, the frequent and sparse labels with respect to the corresponding modality factors are learned to span broader and tighter in the Cosine space, respectively. Extensive experiments over four benchmarks regarding three multi-modal tasks have been conducted, which validates the effectiveness of the proposed loss function on ten baselines.     

Enlightened by the most recent studies on the tackling of the long-tail problem~\cite{long-tail1, long-tail2}, which rely on the training set statistic to work,  our method also leverages such prior to approach the modality bias problem. Though obtained significant performance improvements, this restriction positions our MMDB at the cost of explicit bias conjecture beforehand, limiting its generalization capability across datasets. One potential solution is to discover the biased modalities without presumptions or labels~\cite{bias-discover} in the first step. After that, our method can be seamlessly combined with such techniques to perform with the ease of the prior pain. It is worth mentioning that this work does not present an advanced model for pursuing the SOTA results in multi-modal learning. We instead expect that, with this de-bias loss function, future research can focus more on the enhancement of the multi-modal understanding while with less affect of the modality bias. 

With the recognition of this problem in multi-modal classification, exploring its associated expression for generation tasks will open an interesting door to increase the output diversity. In addition, tackling the modality bias problem from the other two stages, \ie multi-modal representation and fusion, demands our further research attention as well.

%% file: segments/appendix.tex
\appendix \label{appendix}
\section{Proof for the Scaling Factor s} \label{proof}
Without loss of generality, let $P_{i}$ denotes the expected minimum of the class $i$. In the ideal formulation, the $\theta_i$ between label  $i$ and weight vector $\mathbf{W}_i$ should be $0$, while that for others labels should be $180$. We then have:
\begin{equation*}
\begin{aligned}
    \frac{\exp{s (1 - m_{i})}}{\sum_{j \neq i} \exp{s (-1 - m_{j})} + \exp{s (1 - m_{i})}} &\geq P_{i}, \\
    1 + \frac{\sum_{j \neq i} \exp{s (-1 - m_{j})}}{\exp{s (1 - m_{i})}} &\leq \frac{1}{P_{i}}, \\
    1 + \frac{\sum_{j \neq i} \exp{s (m_{j})}}{\exp{s (2 - m_{i})}} &\leq \frac{1}{P_{i}}. \\
\end{aligned}
\end{equation*}
On the basis of Jensen's inequality,
\begin{equation*}
\begin{aligned}
    \sum_{j \neq i} \exp{s (m_{j})} &= \frac{|\Omega| - 1}{|\Omega| - 1} \sum_{j \neq i} \exp{s (m_{j})}, \\
    &\geq (|\Omega| - 1) \exp{s(\frac{\sum_{j \neq i} m_j}{|\Omega| - 1})}.
\end{aligned}
\end{equation*}
Accordingly, we obtain,
\begin{equation*}
\begin{aligned}
    1 + \frac{(|\Omega| - 1) \exp{s(\nicefrac{\sum_{j \neq i} m_j}{(|\Omega| - 1}))}}{\exp{s (2 - m_{i})}} &\leq \frac{1}{P_{i}}, \\
    \exp{s(\frac{\sum_{j \neq i} m_j}{|\Omega| - 1} + m_{i} -2)} &\leq \frac{1}{P_{i}} - 1.\\
\end{aligned}
\end{equation*}
Finally, we have,
\begin{equation} \label{equ:scale_2}
    s \geq  \frac{\ln(\nicefrac{1}{P_{i}} - 1)}{m_{i} +\nicefrac{\sum_{j \neq i} m_j}{(|\Omega| - 1)} - 2}.
\end{equation}

%% file: tomm.bbl

\begin{thebibliography}{66}


\ifx \showCODEN    \undefined \def \showCODEN     #1{\unskip}     \fi
\ifx \showDOI      \undefined \def \showDOI       #1{#1}\fi
\ifx \showISBNx    \undefined \def \showISBNx     #1{\unskip}     \fi
\ifx \showISBNxiii \undefined \def \showISBNxiii  #1{\unskip}     \fi
\ifx \showISSN     \undefined \def \showISSN      #1{\unskip}     \fi
\ifx \showLCCN     \undefined \def \showLCCN      #1{\unskip}     \fi
\ifx \shownote     \undefined \def \shownote      #1{#1}          \fi
\ifx \showarticletitle \undefined \def \showarticletitle #1{#1}   \fi
\ifx \showURL      \undefined \def \showURL       {\relax}        \fi
\providecommand\bibfield[2]{#2}
\providecommand\bibinfo[2]{#2}
\providecommand\natexlab[1]{#1}
\providecommand\showeprint[2][]{arXiv:#2}

\bibitem[Agrawal et~al\mbox{.}(2018)]%
        {vqacp}
\bibfield{author}{\bibinfo{person}{Aishwarya Agrawal}, \bibinfo{person}{Dhruv
  Batra}, \bibinfo{person}{Devi Parikh}, {and} \bibinfo{person}{Aniruddha
  Kembhavi}.} \bibinfo{year}{2018}\natexlab{}.
\newblock \showarticletitle{Don't Just Assume; Look and Answer: Overcoming
  Priors for Visual Question Answering}. In \bibinfo{booktitle}{\emph{{IEEE}
  Conference on Computer Vision and Pattern Recognition}}.
  \bibinfo{publisher}{{IEEE}}, \bibinfo{pages}{4971--4980}.
\newblock


\bibitem[Anderson et~al\mbox{.}(2018)]%
        {updown}
\bibfield{author}{\bibinfo{person}{Peter Anderson}, \bibinfo{person}{Xiaodong
  He}, \bibinfo{person}{Chris Buehler}, \bibinfo{person}{Damien Teney},
  \bibinfo{person}{Mark Johnson}, \bibinfo{person}{Stephen Gould}, {and}
  \bibinfo{person}{Lei Zhang}.} \bibinfo{year}{2018}\natexlab{}.
\newblock \showarticletitle{Bottom-Up and Top-Down Attention for Image
  Captioning and Visual Question Answering}. In
  \bibinfo{booktitle}{\emph{{IEEE} Conference on Computer Vision and Pattern
  Recognition}}. \bibinfo{publisher}{{IEEE}}, \bibinfo{pages}{6077--6086}.
\newblock


\bibitem[Andreas et~al\mbox{.}(2016)]%
        {nmn}
\bibfield{author}{\bibinfo{person}{Jacob Andreas}, \bibinfo{person}{Marcus
  Rohrbach}, \bibinfo{person}{Trevor Darrell}, {and} \bibinfo{person}{Dan
  Klein}.} \bibinfo{year}{2016}\natexlab{}.
\newblock \showarticletitle{Neural Module Networks}. In
  \bibinfo{booktitle}{\emph{{IEEE} Conference on Computer Vision and Pattern
  Recognition}}. \bibinfo{publisher}{{IEEE}}, \bibinfo{pages}{39--48}.
\newblock


\bibitem[Antol et~al\mbox{.}(2015)]%
        {vqa1}
\bibfield{author}{\bibinfo{person}{Stanislaw Antol}, \bibinfo{person}{Aishwarya
  Agrawal}, \bibinfo{person}{Jiasen Lu}, \bibinfo{person}{Margaret Mitchell},
  \bibinfo{person}{Dhruv Batra}, \bibinfo{person}{C.~Lawrence Zitnick}, {and}
  \bibinfo{person}{Devi Parikh}.} \bibinfo{year}{2015}\natexlab{}.
\newblock \showarticletitle{{VQA:} Visual Question Answering}. In
  \bibinfo{booktitle}{\emph{{IEEE} International Conference on Computer
  Vision}}. \bibinfo{publisher}{{IEEE}}, \bibinfo{pages}{2425--2433}.
\newblock


\bibitem[Arnab et~al\mbox{.}(2021)]%
        {vivit}
\bibfield{author}{\bibinfo{person}{Anurag Arnab}, \bibinfo{person}{Mostafa
  Dehghani}, \bibinfo{person}{Georg Heigold}, \bibinfo{person}{Chen Sun},
  \bibinfo{person}{Mario Lucic}, {and} \bibinfo{person}{Cordelia Schmid}.}
  \bibinfo{year}{2021}\natexlab{}.
\newblock \showarticletitle{ViViT: {A} Video Vision Transformer}. In
  \bibinfo{booktitle}{\emph{International Conference on Computer Vision}}.
  \bibinfo{publisher}{{IEEE}}, \bibinfo{pages}{6816--6826}.
\newblock


\bibitem[Bai et~al\mbox{.}(2021a)]%
        {oodfeature}
\bibfield{author}{\bibinfo{person}{Haoyue Bai}, \bibinfo{person}{Rui Sun},
  \bibinfo{person}{Lanqing Hong}, \bibinfo{person}{Fengwei Zhou},
  \bibinfo{person}{Nanyang Ye}, \bibinfo{person}{Han{-}Jia Ye},
  \bibinfo{person}{S.{-}H.~Gary Chan}, {and} \bibinfo{person}{Zhenguo Li}.}
  \bibinfo{year}{2021}\natexlab{a}.
\newblock \showarticletitle{DecAug: Out-of-Distribution Generalization via
  Decomposed Feature Representation and Semantic Augmentation}. In
  \bibinfo{booktitle}{\emph{{AAAI} Conference on Artificial Intelligence}}.
  \bibinfo{publisher}{{AAAI}}, \bibinfo{pages}{6705--6713}.
\newblock


\bibitem[Bai et~al\mbox{.}(2021b)]%
        {ood2}
\bibfield{author}{\bibinfo{person}{Haoyue Bai}, \bibinfo{person}{Fengwei Zhou},
  \bibinfo{person}{Lanqing Hong}, \bibinfo{person}{Nanyang Ye},
  \bibinfo{person}{S-H~Gary Chan}, {and} \bibinfo{person}{Zhenguo Li}.}
  \bibinfo{year}{2021}\natexlab{b}.
\newblock \showarticletitle{NAS-OoD: Neural Architecture Search for
  Out-of-Distribution Generalization}. In \bibinfo{booktitle}{\emph{IEEE/CVF
  International Conference on Computer Vision}}. \bibinfo{publisher}{{IEEE}},
  \bibinfo{pages}{8320--8329}.
\newblock


\bibitem[Bertasius et~al\mbox{.}(2021)]%
        {timesformer}
\bibfield{author}{\bibinfo{person}{Gedas Bertasius}, \bibinfo{person}{Heng
  Wang}, {and} \bibinfo{person}{Lorenzo Torresani}.}
  \bibinfo{year}{2021}\natexlab{}.
\newblock \showarticletitle{Is Space-Time Attention All You Need for Video
  Understanding?}. In \bibinfo{booktitle}{\emph{International Conference on
  Machine Learning}}. \bibinfo{publisher}{{PMLR}}, \bibinfo{pages}{813--824}.
\newblock


\bibitem[Cad{\`{e}}ne et~al\mbox{.}(2019)]%
        {rubi}
\bibfield{author}{\bibinfo{person}{R{\'{e}}mi Cad{\`{e}}ne},
  \bibinfo{person}{Corentin Dancette}, \bibinfo{person}{Hedi Ben{-}younes},
  \bibinfo{person}{Matthieu Cord}, {and} \bibinfo{person}{Devi Parikh}.}
  \bibinfo{year}{2019}\natexlab{}.
\newblock \showarticletitle{RUBi: Reducing Unimodal Biases for Visual Question
  Answering}. In \bibinfo{booktitle}{\emph{Advances in Neural Information
  Processing Systems}}. \bibinfo{publisher}{MIT}, \bibinfo{pages}{839--850}.
\newblock


\bibitem[Caliskan et~al\mbox{.}(2017)]%
        {male_bias}
\bibfield{author}{\bibinfo{person}{Aylin Caliskan}, \bibinfo{person}{Joanna~J
  Bryson}, {and} \bibinfo{person}{Arvind Narayanan}.}
  \bibinfo{year}{2017}\natexlab{}.
\newblock \showarticletitle{Semantics Derived Automatically from Language
  Corpora Contain Human-like Biases}.
\newblock \bibinfo{journal}{\emph{Science}} \bibinfo{volume}{356},
  \bibinfo{number}{6334} (\bibinfo{year}{2017}), \bibinfo{pages}{183--186}.
\newblock


\bibitem[Carreira and Zisserman(2017)]%
        {i3d}
\bibfield{author}{\bibinfo{person}{Jo{\~{a}}o Carreira} {and}
  \bibinfo{person}{Andrew Zisserman}.} \bibinfo{year}{2017}\natexlab{}.
\newblock \showarticletitle{Quo Vadis, Action Recognition? {A} New Model and
  the Kinetics Dataset}. In \bibinfo{booktitle}{\emph{{IEEE} Conference on
  Computer Vision and Pattern Recognition}}. \bibinfo{publisher}{{IEEE}},
  \bibinfo{pages}{4724--4733}.
\newblock


\bibitem[Chen et~al\mbox{.}(2021)]%
        {action}
\bibfield{author}{\bibinfo{person}{Chun{-}Fu~(Richard) Chen},
  \bibinfo{person}{Rameswar Panda}, \bibinfo{person}{Kandan Ramakrishnan},
  \bibinfo{person}{Rog{\'{e}}rio Feris}, \bibinfo{person}{John Cohn},
  \bibinfo{person}{Aude Oliva}, {and} \bibinfo{person}{Quanfu Fan}.}
  \bibinfo{year}{2021}\natexlab{}.
\newblock \showarticletitle{Deep Analysis of CNN-Based Spatio-Temporal
  Representations for Action Recognition}. In \bibinfo{booktitle}{\emph{{IEEE}
  Conference on Computer Vision and Pattern Recognition}}.
  \bibinfo{publisher}{{IEEE}}, \bibinfo{pages}{6165--6175}.
\newblock


\bibitem[Chen et~al\mbox{.}(2020)]%
        {counterfactual}
\bibfield{author}{\bibinfo{person}{Long Chen}, \bibinfo{person}{Xin Yan},
  \bibinfo{person}{Jun Xiao}, \bibinfo{person}{Hanwang Zhang},
  \bibinfo{person}{Shiliang Pu}, {and} \bibinfo{person}{Yueting Zhuang}.}
  \bibinfo{year}{2020}\natexlab{}.
\newblock \showarticletitle{Counterfactual Samples Synthesizing for Robust
  Visual Question Answering}. In \bibinfo{booktitle}{\emph{{IEEE/CVF}
  Conference on Computer Vision and Pattern Recognition}}.
  \bibinfo{publisher}{{IEEE}}, \bibinfo{pages}{10797--10806}.
\newblock


\bibitem[Chen and Joo(2021)]%
        {method_all}
\bibfield{author}{\bibinfo{person}{Yunliang Chen} {and}
  \bibinfo{person}{Jungseock Joo}.} \bibinfo{year}{2021}\natexlab{}.
\newblock \showarticletitle{Understanding and Mitigating Annotation Bias in
  Facial Expression Recognition}. In \bibinfo{booktitle}{\emph{IEEE/CVF
  International Conference on Computer Vision}}. \bibinfo{publisher}{{IEEE}},
  \bibinfo{pages}{14980--14991}.
\newblock


\bibitem[Cheng et~al\mbox{.}(2021)]%
        {nlp_all2}
\bibfield{author}{\bibinfo{person}{Lu Cheng}, \bibinfo{person}{Ahmadreza
  Mosallanezhad}, \bibinfo{person}{Yasin~N. Silva}, \bibinfo{person}{Deborah~L.
  Hall}, {and} \bibinfo{person}{Huan Liu}.} \bibinfo{year}{2021}\natexlab{}.
\newblock \showarticletitle{Mitigating Bias in Session-based Cyberbullying
  Detection: {A} Non-Compromising Approach}. In
  \bibinfo{booktitle}{\emph{Annual Meeting of the Association for Computational
  Linguistics and International Joint Conference on Natural Language
  Processing}}. \bibinfo{publisher}{ACL}, \bibinfo{pages}{2158--2168}.
\newblock


\bibitem[Clark et~al\mbox{.}(2019)]%
        {lmh}
\bibfield{author}{\bibinfo{person}{Christopher Clark}, \bibinfo{person}{Mark
  Yatskar}, {and} \bibinfo{person}{Luke Zettlemoyer}.}
  \bibinfo{year}{2019}\natexlab{}.
\newblock \showarticletitle{Don't Take the Easy Way Out: Ensemble Based Methods
  for Avoiding Known Dataset Biases}. In \bibinfo{booktitle}{\emph{Conference
  on Empirical Methods in Natural Language Processing and International Joint
  Conference on Natural Language Processing}}. \bibinfo{publisher}{ACL},
  \bibinfo{pages}{4067--4080}.
\newblock


\bibitem[Deng et~al\mbox{.}(2019)]%
        {arcface}
\bibfield{author}{\bibinfo{person}{Jiankang Deng}, \bibinfo{person}{Jia Guo},
  \bibinfo{person}{Niannan Xue}, {and} \bibinfo{person}{Stefanos Zafeiriou}.}
  \bibinfo{year}{2019}\natexlab{}.
\newblock \showarticletitle{ArcFace: Additive Angular Margin Loss for Deep Face
  Recognition}. In \bibinfo{booktitle}{\emph{{IEEE} Conference on Computer
  Vision and Pattern Recognition}}. \bibinfo{publisher}{{IEEE}},
  \bibinfo{pages}{4690--4699}.
\newblock


\bibitem[Dixon et~al\mbox{.}(2018)]%
        {external_data1}
\bibfield{author}{\bibinfo{person}{Lucas Dixon}, \bibinfo{person}{John Li},
  \bibinfo{person}{Jeffrey Sorensen}, \bibinfo{person}{Nithum Thain}, {and}
  \bibinfo{person}{Lucy Vasserman}.} \bibinfo{year}{2018}\natexlab{}.
\newblock \showarticletitle{Measuring and Mitigating Unintended Bias in Text
  Classification}. In \bibinfo{booktitle}{\emph{{AAAI/ACM} Conference on AI,
  Ethics, and Society}}. \bibinfo{publisher}{{ACM}}, \bibinfo{pages}{67--73}.
\newblock


\bibitem[Dou et~al\mbox{.}(2019)]%
        {ooddomain}
\bibfield{author}{\bibinfo{person}{Qi Dou}, \bibinfo{person}{Daniel~Coelho de
  Castro}, \bibinfo{person}{Konstantinos Kamnitsas}, {and} \bibinfo{person}{Ben
  Glocker}.} \bibinfo{year}{2019}\natexlab{}.
\newblock \showarticletitle{Domain Generalization via Model-Agnostic Learning
  of Semantic Features}. In \bibinfo{booktitle}{\emph{Advances in Neural
  Information Processing Systems}}. \bibinfo{publisher}{MIT},
  \bibinfo{pages}{6447--6458}.
\newblock


\bibitem[Engstrom et~al\mbox{.}(2020)]%
        {imgnetv2method}
\bibfield{author}{\bibinfo{person}{Logan Engstrom}, \bibinfo{person}{Andrew
  Ilyas}, \bibinfo{person}{Shibani Santurkar}, \bibinfo{person}{Dimitris
  Tsipras}, \bibinfo{person}{Jacob Steinhardt}, {and}
  \bibinfo{person}{Aleksander Madry}.} \bibinfo{year}{2020}\natexlab{}.
\newblock \showarticletitle{Identifying Statistical Bias in Dataset
  Replication}. In \bibinfo{booktitle}{\emph{International Conference on
  Machine Learning}}. \bibinfo{publisher}{{PMLR}}, \bibinfo{pages}{2922--2932}.
\newblock


\bibitem[Fukui et~al\mbox{.}(2016)]%
        {mcb}
\bibfield{author}{\bibinfo{person}{Akira Fukui}, \bibinfo{person}{Dong~Huk
  Park}, \bibinfo{person}{Daylen Yang}, \bibinfo{person}{Anna Rohrbach},
  \bibinfo{person}{Trevor Darrell}, {and} \bibinfo{person}{Marcus Rohrbach}.}
  \bibinfo{year}{2016}\natexlab{}.
\newblock \showarticletitle{Multimodal Compact Bilinear Pooling for Visual
  Question Answering and Visual Grounding}. In
  \bibinfo{booktitle}{\emph{Conference on Empirical Methods in Natural Language
  Processing}}. \bibinfo{publisher}{ACL}, \bibinfo{pages}{457--468}.
\newblock


\bibitem[Gat et~al\mbox{.}(2020)]%
        {mfe}
\bibfield{author}{\bibinfo{person}{Itai Gat}, \bibinfo{person}{Idan Schwartz},
  \bibinfo{person}{Alexander~G. Schwing}, {and} \bibinfo{person}{Tamir Hazan}.}
  \bibinfo{year}{2020}\natexlab{}.
\newblock \showarticletitle{Removing Bias in Multi-modal Classifiers:
  Regularization by Maximizing Functional Entropies}. In
  \bibinfo{booktitle}{\emph{Advances in Neural Information Processing
  Systems}}. \bibinfo{publisher}{MIT}.
\newblock


\bibitem[Gong et~al\mbox{.}(2021)]%
        {racial_bias}
\bibfield{author}{\bibinfo{person}{Sixue Gong}, \bibinfo{person}{Xiaoming Liu},
  {and} \bibinfo{person}{Anil~K. Jain}.} \bibinfo{year}{2021}\natexlab{}.
\newblock \showarticletitle{Mitigating Face Recognition Bias via Group Adaptive
  Classifier}. In \bibinfo{booktitle}{\emph{{IEEE} Conference on Computer
  Vision and Pattern Recognition}}. \bibinfo{publisher}{{IEEE}},
  \bibinfo{pages}{3414--3424}.
\newblock


\bibitem[Goyal et~al\mbox{.}(2017)]%
        {vqa2}
\bibfield{author}{\bibinfo{person}{Yash Goyal}, \bibinfo{person}{Tejas Khot},
  \bibinfo{person}{Douglas Summers{-}Stay}, \bibinfo{person}{Dhruv Batra},
  {and} \bibinfo{person}{Devi Parikh}.} \bibinfo{year}{2017}\natexlab{}.
\newblock \showarticletitle{Making the {V} in {VQA} Matter: Elevating the Role
  of Image Understanding in Visual Question Answering}. In
  \bibinfo{booktitle}{\emph{{IEEE} Conference on Computer Vision and Pattern
  Recognition}}. \bibinfo{publisher}{{IEEE}}, \bibinfo{pages}{6325--6334}.
\newblock


\bibitem[Guo et~al\mbox{.}(2019)]%
        {lpscore}
\bibfield{author}{\bibinfo{person}{Yangyang Guo}, \bibinfo{person}{Zhiyong
  Cheng}, \bibinfo{person}{Liqiang Nie}, \bibinfo{person}{Yibing Liu},
  \bibinfo{person}{Yinglong Wang}, {and} \bibinfo{person}{Mohan~S.
  Kankanhalli}.} \bibinfo{year}{2019}\natexlab{}.
\newblock \showarticletitle{Quantifying and Alleviating the Language Prior
  Problem in Visual Question Answering}. In
  \bibinfo{booktitle}{\emph{International {ACM} {SIGIR} Conference on Research
  and Development in Information Retrieval}}. \bibinfo{publisher}{{ACM}},
  \bibinfo{pages}{75--84}.
\newblock


\bibitem[Guo et~al\mbox{.}(2021a)]%
        {adavqa}
\bibfield{author}{\bibinfo{person}{Yangyang Guo}, \bibinfo{person}{Liqiang
  Nie}, \bibinfo{person}{Zhiyong Cheng}, \bibinfo{person}{Feng Ji},
  \bibinfo{person}{Ji Zhang}, {and} \bibinfo{person}{Alberto~Del Bimbo}.}
  \bibinfo{year}{2021}\natexlab{a}.
\newblock \showarticletitle{AdaVQA: Overcoming Language Priors with Adapted
  Margin Cosine Loss}. In \bibinfo{booktitle}{\emph{International Joint
  Conference on Artificial Intelligence}}. \bibinfo{publisher}{ijcai.org},
  \bibinfo{pages}{708--714}.
\newblock


\bibitem[Guo et~al\mbox{.}(2021b)]%
        {loss-vqa}
\bibfield{author}{\bibinfo{person}{Yangyang Guo}, \bibinfo{person}{Liqiang
  Nie}, \bibinfo{person}{Zhiyong Cheng}, \bibinfo{person}{Qi Tian}, {and}
  \bibinfo{person}{Min Zhang}.} \bibinfo{year}{2021}\natexlab{b}.
\newblock \showarticletitle{Loss Re-scaling VQA: Revisiting the Language Prior
  Problem from A Class-imbalance View}.
\newblock \bibinfo{journal}{\emph{TIP}}.
\newblock


\bibitem[Hama et~al\mbox{.}(2021)]%
        {captiontomm}
\bibfield{author}{\bibinfo{person}{Kenta Hama}, \bibinfo{person}{Takashi
  Matsubara}, \bibinfo{person}{Kuniaki Uehara}, {and} \bibinfo{person}{Jianfei
  Cai}.} \bibinfo{year}{2021}\natexlab{}.
\newblock \showarticletitle{Exploring Uncertainty Measures for Image-caption
  Embedding-and-retrieval Task}.
\newblock \bibinfo{journal}{\emph{{ACM} Trans. Multim. Comput. Commun. Appl.}}
  \bibinfo{volume}{17}, \bibinfo{number}{2} (\bibinfo{year}{2021}),
  \bibinfo{pages}{46:1--46:19}.
\newblock


\bibitem[He et~al\mbox{.}(2016)]%
        {resnet}
\bibfield{author}{\bibinfo{person}{Kaiming He}, \bibinfo{person}{Xiangyu
  Zhang}, \bibinfo{person}{Shaoqing Ren}, {and} \bibinfo{person}{Jian Sun}.}
  \bibinfo{year}{2016}\natexlab{}.
\newblock \showarticletitle{Deep Residual Learning for Image Recognition}. In
  \bibinfo{booktitle}{\emph{{IEEE} Conference on Computer Vision and Pattern
  Recognition}}. \bibinfo{publisher}{{IEEE}}, \bibinfo{pages}{770--778}.
\newblock


\bibitem[Hendrycks et~al\mbox{.}(2021)]%
        {ood1}
\bibfield{author}{\bibinfo{person}{Dan Hendrycks}, \bibinfo{person}{Steven
  Basart}, \bibinfo{person}{Norman Mu}, \bibinfo{person}{Saurav Kadavath},
  \bibinfo{person}{Frank Wang}, \bibinfo{person}{Evan Dorundo},
  \bibinfo{person}{Rahul Desai}, \bibinfo{person}{Tyler Zhu},
  \bibinfo{person}{Samyak Parajuli}, \bibinfo{person}{Mike Guo},
  {et~al\mbox{.}}} \bibinfo{year}{2021}\natexlab{}.
\newblock \showarticletitle{The Many Faces of Robustness: A Critical Analysis
  of Out-of-Distribution Generalization}. In \bibinfo{booktitle}{\emph{IEEE/CVF
  International Conference on Computer Vision}}. \bibinfo{publisher}{{IEEE}},
  \bibinfo{pages}{8340--8349}.
\newblock


\bibitem[Hendrycks et~al\mbox{.}(2019)]%
        {oodpretrain}
\bibfield{author}{\bibinfo{person}{Dan Hendrycks}, \bibinfo{person}{Kimin Lee},
  {and} \bibinfo{person}{Mantas Mazeika}.} \bibinfo{year}{2019}\natexlab{}.
\newblock \showarticletitle{Using Pre-Training Can Improve Model Robustness and
  Uncertainty}. In \bibinfo{booktitle}{\emph{International Conference on
  Machine Learning}}. \bibinfo{publisher}{{PMLR}}, \bibinfo{pages}{2712--2721}.
\newblock


\bibitem[Hinton et~al\mbox{.}(2015)]%
        {kd-hinton}
\bibfield{author}{\bibinfo{person}{Geoffrey~E. Hinton}, \bibinfo{person}{Oriol
  Vinyals}, {and} \bibinfo{person}{Jeffrey Dean}.}
  \bibinfo{year}{2015}\natexlab{}.
\newblock \showarticletitle{Distilling the Knowledge in a Neural Network}.
\newblock \bibinfo{journal}{\emph{CoRR}}  \bibinfo{volume}{abs/1503.02531}
  (\bibinfo{year}{2015}).
\newblock


\bibitem[Huang and Belongie(2017)]%
        {texture_bias2}
\bibfield{author}{\bibinfo{person}{Xun Huang} {and} \bibinfo{person}{Serge~J.
  Belongie}.} \bibinfo{year}{2017}\natexlab{}.
\newblock \showarticletitle{Arbitrary Style Transfer in Real-Time with Adaptive
  Instance Normalization}. In \bibinfo{booktitle}{\emph{{IEEE} International
  Conference on Computer Vision}}. \bibinfo{publisher}{{IEEE}},
  \bibinfo{pages}{1510--1519}.
\newblock


\bibitem[Jing et~al\mbox{.}(2020)]%
        {questiontype}
\bibfield{author}{\bibinfo{person}{Chenchen Jing}, \bibinfo{person}{Yuwei Wu},
  \bibinfo{person}{Xiaoxun Zhang}, \bibinfo{person}{Yunde Jia}, {and}
  \bibinfo{person}{Qi Wu}.} \bibinfo{year}{2020}\natexlab{}.
\newblock \showarticletitle{Overcoming Language Priors in {VQA} via Decomposed
  Linguistic Representations}. In \bibinfo{booktitle}{\emph{{AAAI} Conference
  on Artificial Intelligence}}. \bibinfo{publisher}{{AAAI}},
  \bibinfo{pages}{11181--11188}.
\newblock


\bibitem[Johnson et~al\mbox{.}(2017a)]%
        {clevr}
\bibfield{author}{\bibinfo{person}{Justin Johnson}, \bibinfo{person}{Bharath
  Hariharan}, \bibinfo{person}{Laurens van~der Maaten}, \bibinfo{person}{Li
  Fei{-}Fei}, \bibinfo{person}{C.~Lawrence Zitnick}, {and}
  \bibinfo{person}{Ross~B. Girshick}.} \bibinfo{year}{2017}\natexlab{a}.
\newblock \showarticletitle{{CLEVR:} {A} Diagnostic Dataset for Compositional
  Language and Elementary Visual Reasoning}. In
  \bibinfo{booktitle}{\emph{{IEEE} Conference on Computer Vision and Pattern
  Recognition}}. \bibinfo{publisher}{{IEEE}}, \bibinfo{pages}{1988--1997}.
\newblock


\bibitem[Johnson et~al\mbox{.}(2017b)]%
        {vgqe}
\bibfield{author}{\bibinfo{person}{Justin Johnson}, \bibinfo{person}{Bharath
  Hariharan}, \bibinfo{person}{Laurens van~der Maaten}, \bibinfo{person}{Li
  Fei{-}Fei}, \bibinfo{person}{C.~Lawrence Zitnick}, {and}
  \bibinfo{person}{Ross~B. Girshick}.} \bibinfo{year}{2017}\natexlab{b}.
\newblock \showarticletitle{{CLEVR:} {A} Diagnostic Dataset for Compositional
  Language and Elementary Visual Reasoning}. In
  \bibinfo{booktitle}{\emph{{IEEE} Conference on Computer Vision and Pattern
  Recognition}}. \bibinfo{publisher}{{IEEE}}, \bibinfo{pages}{1988--1997}.
\newblock


\bibitem[Kay et~al\mbox{.}(2017)]%
        {kinetics}
\bibfield{author}{\bibinfo{person}{Will Kay}, \bibinfo{person}{Jo{\~{a}}o
  Carreira}, \bibinfo{person}{Karen Simonyan}, \bibinfo{person}{Brian Zhang},
  \bibinfo{person}{Chloe Hillier}, \bibinfo{person}{Sudheendra
  Vijayanarasimhan}, \bibinfo{person}{Fabio Viola}, \bibinfo{person}{Tim
  Green}, \bibinfo{person}{Trevor Back}, \bibinfo{person}{Paul Natsev},
  \bibinfo{person}{Mustafa Suleyman}, {and} \bibinfo{person}{Andrew
  Zisserman}.} \bibinfo{year}{2017}\natexlab{}.
\newblock \showarticletitle{The Kinetics Human Action Video Dataset}.
\newblock \bibinfo{journal}{\emph{CoRR}}  \bibinfo{volume}{abs/1705.06950}
  (\bibinfo{year}{2017}).
\newblock


\bibitem[Kim et~al\mbox{.}(2021)]%
        {biaswap}
\bibfield{author}{\bibinfo{person}{Eungyeup Kim}, \bibinfo{person}{Jihyeon
  Lee}, {and} \bibinfo{person}{Jaegul Choo}.} \bibinfo{year}{2021}\natexlab{}.
\newblock \showarticletitle{BiaSwap: Removing Dataset Bias with Bias-Tailored
  Swapping Augmentation}. In \bibinfo{booktitle}{\emph{International Conference
  on Computer Vision}}. \bibinfo{publisher}{{IEEE}},
  \bibinfo{pages}{14972--14981}.
\newblock


\bibitem[Kortylewski et~al\mbox{.}(2018)]%
        {3d}
\bibfield{author}{\bibinfo{person}{Adam Kortylewski}, \bibinfo{person}{Bernhard
  Egger}, \bibinfo{person}{Andreas Schneider}, \bibinfo{person}{Thomas Gerig},
  \bibinfo{person}{Andreas Morel{-}Forster}, {and} \bibinfo{person}{Thomas
  Vetter}.} \bibinfo{year}{2018}\natexlab{}.
\newblock \showarticletitle{Empirically Analyzing the Effect of Dataset Biases
  on Deep Face Recognition Systems}. In \bibinfo{booktitle}{\emph{{IEEE}
  Conference on Computer Vision and Pattern Recognition Workshops}}.
  \bibinfo{publisher}{{IEEE}}, \bibinfo{pages}{2093--2102}.
\newblock


\bibitem[LeCun et~al\mbox{.}(1998)]%
        {lenet}
\bibfield{author}{\bibinfo{person}{Yann LeCun}, \bibinfo{person}{L{\'e}on
  Bottou}, \bibinfo{person}{Yoshua Bengio}, {and} \bibinfo{person}{Patrick
  Haffner}.} \bibinfo{year}{1998}\natexlab{}.
\newblock \showarticletitle{Gradient-based Learning Applied to Document
  Recognition}.
\newblock \bibinfo{journal}{\emph{Proc. IEEE}} \bibinfo{volume}{86},
  \bibinfo{number}{11} (\bibinfo{year}{1998}), \bibinfo{pages}{2278--2324}.
\newblock


\bibitem[Lee et~al\mbox{.}(2021)]%
        {long-tail2}
\bibfield{author}{\bibinfo{person}{Hyuck Lee}, \bibinfo{person}{Seungjae Shin},
  {and} \bibinfo{person}{Heeyoung Kim}.} \bibinfo{year}{2021}\natexlab{}.
\newblock \showarticletitle{{ABC:} Auxiliary Balanced Classifier for
  Class-imbalanced Semi-supervised Learning}. In
  \bibinfo{booktitle}{\emph{Advances in Neural Information Processing}}.
  \bibinfo{pages}{7082--7094}.
\newblock


\bibitem[Li et~al\mbox{.}(2021)]%
        {kd}
\bibfield{author}{\bibinfo{person}{Yanchun Li}, \bibinfo{person}{Jianglian
  Cao}, \bibinfo{person}{Zhetao Li}, \bibinfo{person}{Sangyoon Oh}, {and}
  \bibinfo{person}{Nobuyoshi Komuro}.} \bibinfo{year}{2021}\natexlab{}.
\newblock \showarticletitle{Lightweight Single Image Super-resolution with
  Dense Connection Distillation Network}.
\newblock \bibinfo{journal}{\emph{{ACM} Trans. Multim. Comput. Commun. Appl.}}
  \bibinfo{volume}{17}, \bibinfo{number}{1s} (\bibinfo{year}{2021}),
  \bibinfo{pages}{1--17}.
\newblock


\bibitem[Li and Vasconcelos(2019)]%
        {cmnist}
\bibfield{author}{\bibinfo{person}{Yi Li} {and} \bibinfo{person}{Nuno
  Vasconcelos}.} \bibinfo{year}{2019}\natexlab{}.
\newblock \showarticletitle{{REPAIR:} Removing Representation Bias by Dataset
  Resampling}. In \bibinfo{booktitle}{\emph{{IEEE} Conference on Computer
  Vision and Pattern Recognition}}. \bibinfo{publisher}{{IEEE}},
  \bibinfo{pages}{9572--9581}.
\newblock


\bibitem[Li and Xu(2021a)]%
        {age_bias}
\bibfield{author}{\bibinfo{person}{Zhiheng Li} {and} \bibinfo{person}{Chenliang
  Xu}.} \bibinfo{year}{2021}\natexlab{a}.
\newblock \showarticletitle{Discover the Unknown Biased Attribute of An Image
  Classifier}. In \bibinfo{booktitle}{\emph{IEEE/CVF International Conference
  on Computer Vision}}. \bibinfo{publisher}{{IEEE}}.
\newblock


\bibitem[Li and Xu(2021b)]%
        {bias-discover}
\bibfield{author}{\bibinfo{person}{Zhiheng Li} {and} \bibinfo{person}{Chenliang
  Xu}.} \bibinfo{year}{2021}\natexlab{b}.
\newblock \showarticletitle{Discover the Unknown Biased Attribute of an Image
  Classifier}. In \bibinfo{booktitle}{\emph{International Conference on
  Computer Vision}}. \bibinfo{publisher}{{IEEE}},
  \bibinfo{pages}{14950--14959}.
\newblock


\bibitem[Lu et~al\mbox{.}(2019)]%
        {gender_bias}
\bibfield{author}{\bibinfo{person}{Boyu Lu}, \bibinfo{person}{Jun{-}Cheng
  Chen}, \bibinfo{person}{Carlos~Domingo Castillo}, {and} \bibinfo{person}{Rama
  Chellappa}.} \bibinfo{year}{2019}\natexlab{}.
\newblock \showarticletitle{An Experimental Evaluation of Covariates Effects on
  Unconstrained Face Verification}.
\newblock \bibinfo{journal}{\emph{{IEEE} Trans. Biom. Behav. Identity Sci.}}
  \bibinfo{volume}{1}, \bibinfo{number}{1} (\bibinfo{year}{2019}),
  \bibinfo{pages}{42--55}.
\newblock


\bibitem[Noori et~al\mbox{.}(2020)]%
        {concate}
\bibfield{author}{\bibinfo{person}{Farzan~Majeed Noori},
  \bibinfo{person}{Michael Riegler}, \bibinfo{person}{Md.~Zia Uddin}, {and}
  \bibinfo{person}{Jim T{\o}rresen}.} \bibinfo{year}{2020}\natexlab{}.
\newblock \showarticletitle{Human Activity Recognition from Multiple Sensors
  Data Using Multi-fusion Representations and CNNs}.
\newblock \bibinfo{journal}{\emph{{ACM} Trans. Multim. Comput. Commun. Appl.}}
  \bibinfo{volume}{16}, \bibinfo{number}{2} (\bibinfo{year}{2020}),
  \bibinfo{pages}{45:1--45:19}.
\newblock


\bibitem[Park et~al\mbox{.}(2018)]%
        {external_data2}
\bibfield{author}{\bibinfo{person}{Ji~Ho Park}, \bibinfo{person}{Jamin Shin},
  {and} \bibinfo{person}{Pascale Fung}.} \bibinfo{year}{2018}\natexlab{}.
\newblock \showarticletitle{Reducing Gender Bias in Abusive Language
  Detection}. In \bibinfo{booktitle}{\emph{Conference on Empirical Methods in
  Natural Language Processing}}. \bibinfo{publisher}{ACL},
  \bibinfo{pages}{2799--2804}.
\newblock


\bibitem[Perez et~al\mbox{.}(2022)]%
        {redbias}
\bibfield{author}{\bibinfo{person}{Ethan Perez}, \bibinfo{person}{Saffron
  Huang}, \bibinfo{person}{H.~Francis Song}, \bibinfo{person}{Trevor Cai},
  \bibinfo{person}{Roman Ring}, \bibinfo{person}{John Aslanides},
  \bibinfo{person}{Amelia Glaese}, \bibinfo{person}{Nat McAleese}, {and}
  \bibinfo{person}{Geoffrey Irving}.} \bibinfo{year}{2022}\natexlab{}.
\newblock \showarticletitle{Red Teaming Language Models with Language Models}.
  In \bibinfo{booktitle}{\emph{CoRR}}, Vol.~\bibinfo{volume}{abs/2202.03286}.
\newblock


\bibitem[Perez et~al\mbox{.}(2018)]%
        {film}
\bibfield{author}{\bibinfo{person}{Ethan Perez}, \bibinfo{person}{Florian
  Strub}, \bibinfo{person}{Harm de Vries}, \bibinfo{person}{Vincent Dumoulin},
  {and} \bibinfo{person}{Aaron~C. Courville}.} \bibinfo{year}{2018}\natexlab{}.
\newblock \showarticletitle{FiLM: Visual Reasoning with a General Conditioning
  Layer}. In \bibinfo{booktitle}{\emph{Proceedings of the Thirty-Second {AAAI}
  Conference on Artificial Intelligence}}. \bibinfo{publisher}{{AAAI}},
  \bibinfo{pages}{3942--3951}.
\newblock


\bibitem[Pernici et~al\mbox{.}(2019)]%
        {alberto}
\bibfield{author}{\bibinfo{person}{Federico Pernici}, \bibinfo{person}{Matteo
  Bruni}, \bibinfo{person}{Claudio Baecchi}, {and} \bibinfo{person}{Alberto~Del
  Bimbo}.} \bibinfo{year}{2019}\natexlab{}.
\newblock \showarticletitle{Maximally Compact and Separated Features with
  Regular Polytope Networks}. In \bibinfo{booktitle}{\emph{Computer Vision and
  Pattern Recognition Workshops}}. \bibinfo{publisher}{{IEEE}},
  \bibinfo{pages}{46--53}.
\newblock


\bibitem[Ramakrishnan et~al\mbox{.}(2018)]%
        {adversarial-nips}
\bibfield{author}{\bibinfo{person}{Sainandan Ramakrishnan},
  \bibinfo{person}{Aishwarya Agrawal}, {and} \bibinfo{person}{Stefan Lee}.}
  \bibinfo{year}{2018}\natexlab{}.
\newblock \showarticletitle{Overcoming Language Priors in Visual Question
  Answering with Adversarial Regularization}. In
  \bibinfo{booktitle}{\emph{Advances in Neural Information Processing
  Systems}}. \bibinfo{publisher}{MIT}, \bibinfo{pages}{1548--1558}.
\newblock


\bibitem[Recht et~al\mbox{.}(2019)]%
        {imgnetv2}
\bibfield{author}{\bibinfo{person}{Benjamin Recht}, \bibinfo{person}{Rebecca
  Roelofs}, \bibinfo{person}{Ludwig Schmidt}, {and} \bibinfo{person}{Vaishaal
  Shankar}.} \bibinfo{year}{2019}\natexlab{}.
\newblock \showarticletitle{Do ImageNet Classifiers Generalize to ImageNet?}.
  In \bibinfo{booktitle}{\emph{International Conference on Machine Learning}}.
  \bibinfo{publisher}{{PMLR}}, \bibinfo{pages}{5389--5400}.
\newblock


\bibitem[Selvaraju et~al\mbox{.}(2019)]%
        {hint}
\bibfield{author}{\bibinfo{person}{Ramprasaath~Ramasamy Selvaraju},
  \bibinfo{person}{Stefan Lee}, \bibinfo{person}{Yilin Shen},
  \bibinfo{person}{Hongxia Jin}, \bibinfo{person}{Shalini Ghosh},
  \bibinfo{person}{Larry~P. Heck}, \bibinfo{person}{Dhruv Batra}, {and}
  \bibinfo{person}{Devi Parikh}.} \bibinfo{year}{2019}\natexlab{}.
\newblock \showarticletitle{Taking a {HINT:} Leveraging Explanations to Make
  Vision and Language Models More Grounded}. In
  \bibinfo{booktitle}{\emph{{IEEE/CVF} International Conference on Computer
  Vision}}. \bibinfo{publisher}{{IEEE}}, \bibinfo{pages}{2591--2600}.
\newblock


\bibitem[Shah et~al\mbox{.}(2020)]%
        {nlp_all1}
\bibfield{author}{\bibinfo{person}{Deven Shah}, \bibinfo{person}{H.~Andrew
  Schwartz}, {and} \bibinfo{person}{Dirk Hovy}.}
  \bibinfo{year}{2020}\natexlab{}.
\newblock \showarticletitle{Predictive Biases in Natural Language Processing
  Models: {A} Conceptual Framework and Overview}. In
  \bibinfo{booktitle}{\emph{Annual Meeting of the Association for Computational
  Linguistics}}. \bibinfo{publisher}{ACL}, \bibinfo{pages}{5248--5264}.
\newblock


\bibitem[Tan and Bansal(2019)]%
        {lxmert}
\bibfield{author}{\bibinfo{person}{Hao Tan} {and} \bibinfo{person}{Mohit
  Bansal}.} \bibinfo{year}{2019}\natexlab{}.
\newblock \showarticletitle{{LXMERT:} Learning Cross-Modality Encoder
  Representations from Transformers}. In \bibinfo{booktitle}{\emph{Conference
  on Empirical Methods in Natural Language Processing and International Joint
  Conference on Natural Language Processing}}. \bibinfo{publisher}{ACL},
  \bibinfo{pages}{5099--5110}.
\newblock


\bibitem[Teney et~al\mbox{.}(2021)]%
        {unshuffle}
\bibfield{author}{\bibinfo{person}{Damien Teney}, \bibinfo{person}{Ehsan
  Abbasnejad}, {and} \bibinfo{person}{Anton van~den Hengel}.}
  \bibinfo{year}{2021}\natexlab{}.
\newblock \showarticletitle{Unshuffling Data for Improved Generalization in
  Visual Question Answering}. In \bibinfo{booktitle}{\emph{IEEE/CVF
  International Conference on Computer Vision}}. \bibinfo{publisher}{{IEEE}},
  \bibinfo{pages}{1417--1427}.
\newblock


\bibitem[Wang et~al\mbox{.}(2019b)]%
        {texture_bias1}
\bibfield{author}{\bibinfo{person}{Haohan Wang}, \bibinfo{person}{Zexue He},
  \bibinfo{person}{Zachary~C. Lipton}, {and} \bibinfo{person}{Eric~P. Xing}.}
  \bibinfo{year}{2019}\natexlab{b}.
\newblock \showarticletitle{Learning Robust Representations by Projecting
  Superficial Statistics Out}. In \bibinfo{booktitle}{\emph{International
  Conference on Learning Representations}}.
  \bibinfo{publisher}{OpenReview.net}.
\newblock


\bibitem[Wang et~al\mbox{.}(2018)]%
        {cosface}
\bibfield{author}{\bibinfo{person}{Hao Wang}, \bibinfo{person}{Yitong Wang},
  \bibinfo{person}{Zheng Zhou}, \bibinfo{person}{Xing Ji},
  \bibinfo{person}{Dihong Gong}, \bibinfo{person}{Jingchao Zhou},
  \bibinfo{person}{Zhifeng Li}, {and} \bibinfo{person}{Wei Liu}.}
  \bibinfo{year}{2018}\natexlab{}.
\newblock \showarticletitle{CosFace: Large Margin Cosine Loss for Deep Face
  Recognition}. In \bibinfo{booktitle}{\emph{{IEEE} Conference on Computer
  Vision and Pattern Recognition}}. \bibinfo{publisher}{{IEEE}},
  \bibinfo{pages}{5265--5274}.
\newblock


\bibitem[Wang et~al\mbox{.}(2019a)]%
        {domain}
\bibfield{author}{\bibinfo{person}{Mei Wang}, \bibinfo{person}{Weihong Deng},
  \bibinfo{person}{Jiani Hu}, \bibinfo{person}{Xunqiang Tao}, {and}
  \bibinfo{person}{Yaohai Huang}.} \bibinfo{year}{2019}\natexlab{a}.
\newblock \showarticletitle{Racial Faces in the Wild: Reducing Racial Bias by
  Information Maximization Adaptation Network}. In
  \bibinfo{booktitle}{\emph{{IEEE/CVF} International Conference on Computer
  Vision}}. \bibinfo{publisher}{{IEEE}}, \bibinfo{pages}{692--702}.
\newblock


\bibitem[Wang et~al\mbox{.}(2019c)]%
        {adversarial}
\bibfield{author}{\bibinfo{person}{Tianlu Wang}, \bibinfo{person}{Jieyu Zhao},
  \bibinfo{person}{Mark Yatskar}, \bibinfo{person}{Kai{-}Wei Chang}, {and}
  \bibinfo{person}{Vicente Ordonez}.} \bibinfo{year}{2019}\natexlab{c}.
\newblock \showarticletitle{Balanced Datasets Are Not Enough: Estimating and
  Mitigating Gender Bias in Deep Image Representations}. In
  \bibinfo{booktitle}{\emph{{IEEE/CVF} International Conference on Computer
  Vision}}. \bibinfo{publisher}{{IEEE}}, \bibinfo{pages}{5309--5318}.
\newblock


\bibitem[Wu and Mooney(2019)]%
        {critical}
\bibfield{author}{\bibinfo{person}{Jialin Wu} {and} \bibinfo{person}{Raymond~J.
  Mooney}.} \bibinfo{year}{2019}\natexlab{}.
\newblock \showarticletitle{Self-Critical Reasoning for Robust Visual Question
  Answering}. In \bibinfo{booktitle}{\emph{Advances in Neural Information
  Processing Systems}}. \bibinfo{publisher}{MIT}, \bibinfo{pages}{8601--8611}.
\newblock


\bibitem[Zhang et~al\mbox{.}(2020)]%
        {external_data3}
\bibfield{author}{\bibinfo{person}{Guanhua Zhang}, \bibinfo{person}{Bing Bai},
  \bibinfo{person}{Junqi Zhang}, \bibinfo{person}{Kun Bai},
  \bibinfo{person}{Conghui Zhu}, {and} \bibinfo{person}{Tiejun Zhao}.}
  \bibinfo{year}{2020}\natexlab{}.
\newblock \showarticletitle{Demographics Should Not Be the Reason of Toxicity:
  Mitigating Discrimination in Text Classifications with Instance Weighting}.
  In \bibinfo{booktitle}{\emph{Annual Meeting of the Association for
  Computational Linguistics}}. \bibinfo{publisher}{ACL},
  \bibinfo{pages}{4134--4145}.
\newblock


\bibitem[Zhang et~al\mbox{.}(2018)]%
        {counter}
\bibfield{author}{\bibinfo{person}{Yan Zhang}, \bibinfo{person}{Jonathon~S.
  Hare}, {and} \bibinfo{person}{Adam Pr{\"{u}}gel{-}Bennett}.}
  \bibinfo{year}{2018}\natexlab{}.
\newblock \showarticletitle{Learning to Count Objects in Natural Images for
  Visual Question Answering}. In \bibinfo{booktitle}{\emph{International
  Conference on Learning Representations}}.
  \bibinfo{publisher}{OpenReview.net}.
\newblock


\bibitem[Zhang et~al\mbox{.}(2021)]%
        {long-tail1}
\bibfield{author}{\bibinfo{person}{Yongshun Zhang}, \bibinfo{person}{Xiu{-}Shen
  Wei}, \bibinfo{person}{Boyan Zhou}, {and} \bibinfo{person}{Jianxin Wu}.}
  \bibinfo{year}{2021}\natexlab{}.
\newblock \showarticletitle{Bag of Tricks for Long-Tailed Visual Recognition
  with Deep Convolutional Neural Networks}. In
  \bibinfo{booktitle}{\emph{Thirty-Fifth {AAAI} Conference on Artificial
  Intelligence}}. \bibinfo{publisher}{{AAAI}}, \bibinfo{pages}{3447--3455}.
\newblock


\bibitem[Zhuang et~al\mbox{.}(2020)]%
        {addition}
\bibfield{author}{\bibinfo{person}{Yueting Zhuang}, \bibinfo{person}{Dejing
  Xu}, \bibinfo{person}{Xin Yan}, \bibinfo{person}{Wenzhuo Cheng},
  \bibinfo{person}{Zhou Zhao}, \bibinfo{person}{Shiliang Pu}, {and}
  \bibinfo{person}{Jun Xiao}.} \bibinfo{year}{2020}\natexlab{}.
\newblock \showarticletitle{Multichannel Attention Refinement for Video
  Question Answering}.
\newblock \bibinfo{journal}{\emph{{ACM} Trans. Multim. Comput. Commun. Appl.}}
  \bibinfo{volume}{16}, \bibinfo{number}{1s} (\bibinfo{year}{2020}),
  \bibinfo{pages}{1--23}.
\newblock


\end{thebibliography}
